\documentclass[10pt,twocolumn,letterpaper]{article}

\usepackage[pagenumbers]{wacv} 

\usepackage{graphicx}
\usepackage{amsmath}
\usepackage{amssymb}
\usepackage{booktabs}
\usepackage{multirow}
\usepackage{colortbl} 
\usepackage{xcolor} 

\usepackage[pagebackref,breaklinks,colorlinks]{hyperref}

\usepackage[capitalize]{cleveref}
\crefname{section}{Sec.}{Secs.}
\Crefname{section}{Section}{Sections}
\Crefname{table}{Table}{Tables}
\crefname{table}{Tab.}{Tabs.}

\def\wacvPaperID{848} 
\def\confName{WACV}
\def\confYear{2025}

\begin{document}
\definecolor{LightRed}{HTML}{ffe4e6} 
\definecolor{LightBlue}{HTML}{cffafe}
\definecolor{LightYellow}{HTML}{ffedd5}
\definecolor{LightPurple}{HTML}{f3e8ff}
\definecolor{LightGray}{HTML}{e5e7eb}
\title{SpiralMLP: A Lightweight Vision MLP Architecture}

\author{
Haojie Mu \quad Burhan Ul Tayyab \quad Nicholas Chua \\
Kookree\\
{\tt\small \{mu, burhan, nicholas\}@kookee.ai}
\and
}

\maketitle

\begin{abstract}
We present $\textbf{SpiralMLP}$, a novel architecture introduces a $\textbf{Spiral FC}$ layer as a replacement for the conventional Token Mixing approach. Differing from several existing MLP-based models that primarily emphasize axes, our Spiral FC layer is designed as a deformable convolution layer with spiral-like offsets. We further adapt Spiral FC into two variants: $\textbf{Self-Spiral FC}$ and $\textbf{Cross-Spiral FC}$, enabling both local and global feature integration seamlessly, eliminating the need for additional processing steps. To thoroughly investigate the effectiveness of the spiral-like offsets and validate our design, we conduct ablation studies and explore optimal configurations. In empirical tests, SpiralMLP reaches state-of-the-art performance, similar to Transformers, CNNs, and other MLPs, benchmarking on ImageNet-1k, COCO and ADE20K. SpiralMLP still maintains linear computational complexity $O(HW)$ and is compatible with varying input image resolutions. Our study reveals that targeting the full receptive field is not essential for achieving high performance, instead, adopting a refined approach offers better results.
\end{abstract}


\section{Introduction}
\label{sec:intro}
\subsection{Background}
Earlier image classification systems mainly relied on CNN-based architectures~\cite{he2015deep, zoph2018learning, simonyan2015deep, tan2020efficientnet}, which excel with controlled datasets but struggle with biased or uncontrolled conditions. Subsequently, Transformer-based architectures~\cite{vaswani2023attention, dosovitskiy2021image, kolesnikov2020big, bao2022beit} have emerged as alternatives, benefiting from self-attention mechanism that excel with large datasets~\cite{sun2017revisiting} and are adaptable for various tasks~\cite{liu2021swin}. However, they are often more expensive in pretraining and need specific datasets for better performance on downstream tasks.

MLP-based architectures~\cite{tolstikhin2021mlpmixer, liu2021pay} have also shown promise in computer vision tasks, matching Transformer performance with a more data-efficient and lighter design. These systems use two main components: \textbf{Channel Mixing}, which projects features along the channel dimension, and \textbf{Token Mixing}, which captures spatial information by projecting feature along the spatial dimension. These mixing layers collectively enhance context aggregation, improving robustness and reducing training resource needs.

\begin{figure}[tb]
  \centering
  \fbox{\includegraphics[width=0.98\columnwidth]{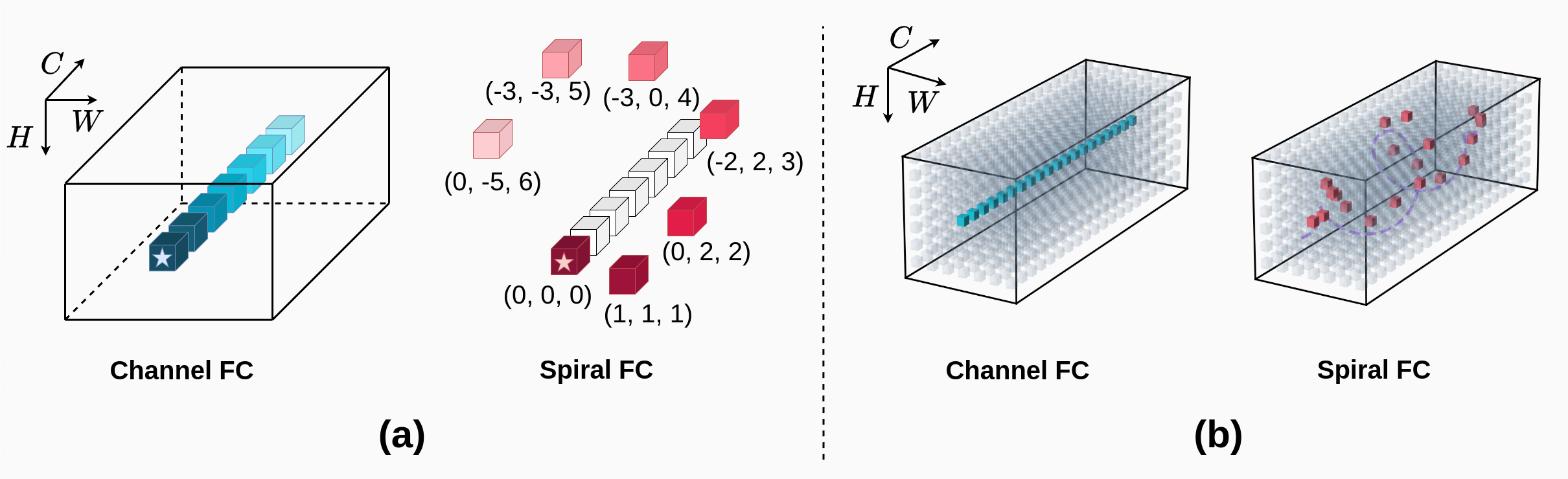}}
   \caption{\textbf{(a)} While the Channel FC concentrates solely at the target point, marked with a $\bigstar$, the Spiral FC captures richer spatial information. Spiral FC is in accordance with~\cref{eq:6,eq:7}, the input channel dimension $C_{\text{in}}=14$, the maximum amplitude $A_{\text{max}}=6$ and $T=8$. The coordinate numbers are arranged as $(H, W, C)$. This illustrative example only contains half of the $C_{\text{in}}$. \textbf{(b)} provides a complete visualization when the parameters are: $C_{\text{in}}=20$, $A_{\text{max}}=3$ and $T = 8$.}
   \label{fig:1}
\end{figure}

\begin{figure*}[tb]
  \centering
    \fbox{\includegraphics[width=0.95\linewidth]{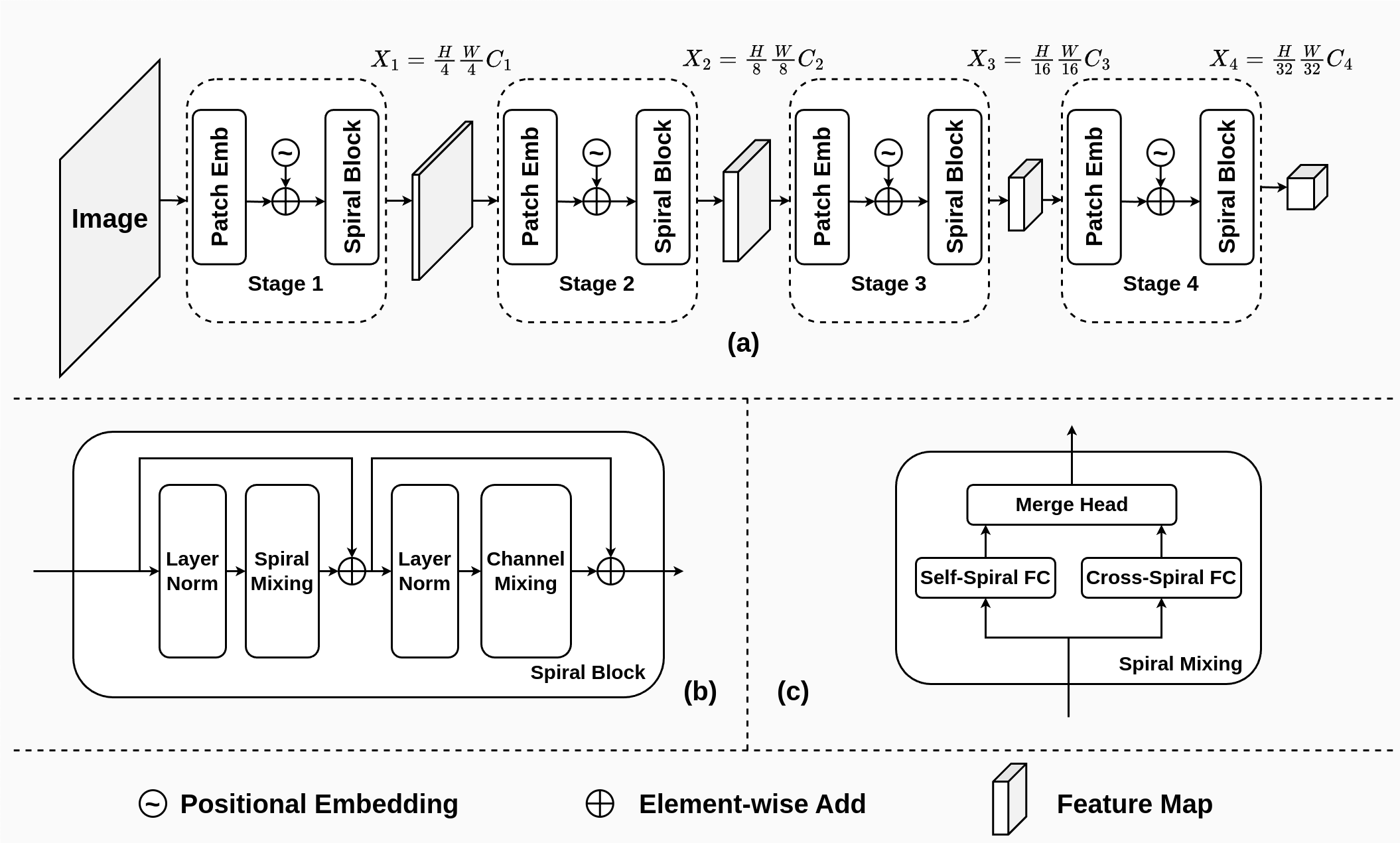}}
    \caption{\textbf{(a)} displays the comprehensive architecture of SpiralMLP in PVT-style, featuring four distinct stages. Each stage is composed of multiple Patch Embedding layers and identically-configured Spiral Blocks. \textbf{(b)} explores the internal layout of a Spiral Block, where the proposed Spiral Mixing replaces the traditional Token Mixing. \textbf{(c)} outlines the components of Spiral Mixing, which incorporates the meticulously designed Spiral FC to effectively capture spatial information. }
    \label{fig:2}
\end{figure*}

\subsection{MLP-Based Architectures. }
The pioneering MLP-Mixer~\cite{tolstikhin2021mlpmixer} proposes a simple yet powerful architecture with both \textbf{Token Mixing} and \textbf{Channel Mixing}. Given a feature map $X\in\mathbb{R}^{H\times W\times C_{\text{in}}}$, where $H$, $W$ are the height and weight, $C_{\text{in}}$ is the input channel dimension, let $W^{\text{Tmix}}\in\mathbb{R}^{H\cdot W\times H\cdot W}$ denote the token mixing weight matrix, the operation applied to the reshaped input $X^T\in\mathbb{R}^{C_{\text{in}}\times H\cdot W}$ is described as follows: 
\begin{equation}
    \text{Tmix}(X)=(X^TW^{\text{Tmix}})^T\label{eq:3}
\end{equation}

where, $\mathbb{R}^{H\cdot W}$ indicates the dimensions are flattened while $\mathbb{R}^{H\times W}$ denotes the dimensions are separated, and $\text{Tmix}(\cdot)\in\mathbb{R}^{H\cdot W\times C_{\text{in}}}$ is the output of token mixing.~\cref{eq:3} is to simulate the attention operation to integrate spatial information, it is followed by the channel mixing that operates along the channel dimension. We define the channel mixing weight matrix as $W^{\text{Cmix}}\in\mathbb{R}^{C_{\text{in}}\times C_{\text{out}}}$, the channel mixing output $\text{Cmix}(\cdot)\in\mathbb{R}^{H\times W\times C_{\text{out}}}$ is expressed as follows: 
\begin{equation}
    \text{Cmix}(\text{Tmix}(X))=(X^TW^{\text{Tmix}})^TW^{\text{Cmix}}\label{eq:4}
\end{equation}

While MLP-Mixer shows strong performance, it is limited by its quadratic computational complexity $O(H^2W^2)$ (\cref{eq:3}) and requires fixed image sizes due to its fully-connected token mixing layer. Alternatives like gMLP~\cite{liu2021pay} introduces a spatial gating unit for better integration, FNet~\cite{leethorp2022fnet} uses Fourier transforms for token mixing, and HireMLP~\cite{guo2021hiremlp} mimics self-attention by swapping elements across regions. Other developments include ResMLP~\cite{touvron2021resmlp}, which replaces LayerNorm with trainable matrices, s$^2$MLP~\cite{yu2021s2mlp} utilizes shifting operations, WaveMLP~\cite{tang2022image} treats pixels as complex numbers, ViP~\cite{hou2021vision} employs a permutator for spatial data, and MorphMLP~\cite{zhang2022morphmlp} gradually expands its receptive field.

Despite advancements, the MLPs mentioned earlier haven't significantly cut computational complexity, leaving an opening for SparseMLP~\cite{tang2022sparse}, ASMLP~\cite{lian2022asmlp}, CycleMLP~\cite{chen2022cyclemlp}, and ATM~\cite{wei2022active}. SparseMLP and ASMLP use dense token mixing along the channel dimension, while CycleMLP introduces Cycle FC for sparser, channel-wise mixing with fixed offsets $S_H$ and $S_W$. ATM~\cite{wei2022active}, on the other hand, uses trainable offsets for dynamic token mixing. However, these models restrict token mixing to horizontal and vertical axes, limiting their ability to integrate feature information across different spatial dimensions.

To address these challenges, we present \textbf{SpiralMLP} with its core component, \textbf{Spiral FC}, based on Channel FC as shown in~\cref{fig:1}(a). Spiral FC offers sufficient receptive field coverage while maintaining linear computational complexity. The paper is organized as follows:
\begin{itemize}
    \item We introduce the SpiralMLP architecture and its foundational Spiral FC layer. 
    \item We conduct experiments to demonstrate SpiralMLP's superiority over other state-of-the-art models.
    \item We perform ablation studies to explore optimal configurations, followed by conclusions and discussions on potential future improvements.
\end{itemize}


\section{Methodology}
\label{sec:methodology}
\subsection{Spiral FC}
We aim to design a compact token mixing layer that captures spatial information efficiently. Our review indicates that traditional designs with criss-cross fully-connected layers fail to optimize the offset function, resulting in inadequate spatial coverage. To address these challenges, we draw inspiration from natural spiral patterns and AttentionViz~\cite{yeh2023attentionviz}, noted for its spiral patterns in transformer attention visualizations.

As a result, we introduce the \textbf{Spiral Fully-Connected Layer} (\textbf{Spiral FC}), intended to replace standard Token Mixing (\cref{eq:3}) in the MLP-Mixer architecture. Described in~\cref{fig:1}(b), Spiral FC leverages a \textbf{spiral trajectory} across the feature map $X\in\mathbb{R}^{H\times W\times C_{\text{in}}}$: 

\begin{equation}
    \text{Spiral FC}_{i,j,:}
(X) = \sum^{C_{\text{in}}}_{c=0}X_{i+\phi_i(c),j+\phi_j(c),c}W^{\text{spiral}}_{c,:}+b^{\text{spiral}}\label{eq:5}
\end{equation}

where, $W^{\text{spiral}}\in\mathbb{R}^{C_{\text{in}}\times C_{\text{out}}}$, $b^{\text{spiral}}\in\mathbb{R}^{C_{\text{out}}}$ are the trainable matrix and bias, $\text{Spiral FC}_{i,j,:}(\cdot)$ is the output at position $(i,j,:)$. Both $\phi_i(c)$ and $\phi_j(c)$ serve as the offset functions along $H,W$ axes respectively within $X$. Furthermore, with the central axis of the spiral trajectory aligns along the channel dimension, the offset functions $\phi_i(c)$ and $\phi_j(c)$ are defined in a spiral manner:  
\begin{align}
    \phi_i(c) = A(c)\cos{(\frac{c\times2\pi}{T})}\label{eq:6}\\
    \phi_j(c) = A(c)\sin{(\frac{c\times2\pi}{T})}\label{eq:7}
\end{align}

where, $T$ is the constant period, $A(\cdot)$ is the amplitude that controls the width of the spiral trajectory, for conciseness, we formulate the amplitude function $A(\cdot)$ with the basic pattern~\footnote{~\cref{sec:abn:update} provides additional cases with universal offset functions.}:
\begin{equation}
    A(c) = \begin{cases}
    \lfloor\frac{2A_{\text{max}}}{C_{\text{in}}}c\rfloor, &0\le c<\frac{C_{\text{in}}}{2} \\
    \lfloor2A_{\text{max}}-\frac{2A_{\text{max}}}{C_{\text{in}}}c\rfloor, &\frac{C_{\text{in}}}{2}\le c\le C_{\text{in}}
\end{cases}\label{eq:8}
\end{equation}

where, $A_{\text{max}}$ is the maximum amplitude. When $A_{\text{max}}=0$, the Spiral FC is identical to Channel FC, denoted as \textbf{Self-Spiral FC}. Conversely, when $A_{\text{max}}\ne0$, it is termed as \textbf{Cross-Spiral FC}. Additionally, we employ a sliding window with a stepsize of $1$. It not only makes the Spiral FC agnostic to the input size, but also enables the flexible feature extraction through meticulously modifying the offset functions (\cref{eq:6,eq:7}), thereby ensuring the Spiral FC operate with linear computational complexity.

\subsection{Spiral Mixing}
At a specific position $(i,j,:)$, Self-Spiral FC captures the local information from itself, yielding an output denoted as $X^{\text{self}}_{i,j,:}$. Conversely, Cross-Spiral FC selectively incorporates spatial information from within the receptive field which is determined by $A_{\text{max}}$, and the output is represented as $X^{\text{cross}}_{i,j,:}$. Across the whole feature map, both the Self-Spiral FC and Cross-Spiral FC operate in parallel, and their outputs, $X^{\text{self}}\in\mathbb{R}^{H\times W\times C_{\text{out}}}$ and $X^{\text{cross}}\in\mathbb{R}^{H\times W\times C_{\text{out}}}$, merge together in the subsequent \textbf{Merge Head}\footnote{Detailed explanation is provided in Appendix.}: 
\begin{equation}
        a=\sigma(W^{\text{merge}}\times[\frac{1}{HW}\sum^{HW}_{i=0}\mathcal{F}(X^{\text{self}}+X^{\text{cross}})_{i,:}])\label{eq:9}
\end{equation}
    
where, the reshaping function $\mathcal{F} : \mathbb{R}^{H \times W \times C_{\text{out}}} \rightarrow \mathbb{R}^{H\cdot W \times C_{\text{out}}}$ flattens the first two dimensions of the input, creating a new projection along the $HW$ dimension. Then, the newly generated projection is averaged into $\mathbb{R}^{1\times C_{\text{out}}}$. Subsequently, $W^{\text{merge}} \in \mathbb{R}^{2,1}$ maps this average from $\mathbb{R}^{1 \times C_{\text{out}}}$ to $\mathbb{R}^{2 \times C_{\text{out}}}$. Finally, the SoftMax function $\sigma(\cdot)$ determines the weights $a\in\mathbb{R}^{2\times C_{\text{out}}}$. Then at position $(i,j,:)$, the Merge Head generates the output: 

\begin{equation}
    X^{\text{spiral}}_{i,j,:}=a_{1,:}\odot X^{\text{self}}_{i,j,:} + a_{2,:}\odot X^{\text{cross}}_{i,j,:}\label{eq:10}
\end{equation}

where, $\odot$ represents the element-wise multiplication. The weights $a$ is to modulate the contribution of the inputs. Furthermore, across the entire $X^{\text{spiral}}$, the weights $a$ is broadcast to influence all elements in both $X^{\text{self}}$ and $X^{\text{cross}}$.

Collectively, Self-Spiral FC, Cross-Spiral FC and Merge Head together constitute the \textbf{Spiral Mixing}, as depicted in~\cref{fig:2} (c). Spiral Mixing transforms the input feature map $X\in\mathbb{R}^{H\times W\times C_{\text{in}}}$ to $X^{\text{spiral}}\in\mathbb{R}^{H\times W\times C_{\text{out}}}$, functioning similarly to vanilla Token Mixing.

\subsection{Spiral Block}

The output $X^{\text{spiral}}$ of Spiral Mixing subsequently proceeds to the \textbf{Channel Mixing} structured as a MLP with a GeLU~\cite{hendrycks2023gaussian} activation function $\zeta(\cdot)$: 
\begin{equation}
    X^{\text{chn}}=\zeta(X^{\text{spiral}}\times W^{\text{mlp1}})\times W^{\text{mlp2}}\label{eq:11}
\end{equation}

where, $W^{\text{mlp1}}\in\mathbb{R}^{C_{\text{out}}\times C_{\text{mlp}}}$ and $W^{\text{mlp2}}\in\mathbb{R}^{C_{\text{mlp}}\times C_{\text{out}}}$ are the linear layer weight matrices. $X^{\text{chn}}$ is the output of Channel Mixing.

Spiral Mixing and Channel Mixing collectively compose the \textbf{Spiral Block}, as depicted in~\cref{fig:2} (b). To summarize, Spiral Block accepts the feature map $X\in\mathbb{R}^{H\times W\times C_{\text{in}}}$ as the input, and initially processes it through a LayerNorm~\cite{ba2016layer} before the Spiral Mixing. Then it produces $X^\prime$ integrated with a residual connection. Following this, $X^\prime$ is processed through another LayerNorm and then Channel Mixing, coupled with another residual connection, resulting in the output $Y$:

\begin{align}
    X^\prime&=\text{Spiral Mixing}(\text{LN}(X))+X\label{eq:12}\\
    Y&=\text{Channel Mixing}(\text{LN}(X'))+X^\prime\label{eq:13}
\end{align}

\subsection{Overall Architecture and Model Zoo}

We firstly construct our \textbf{SpiralMLP} based on the PVT~\cite{wang2022pvt} framework, the models are scaled from \textbf{SpiralMLP-B1} to \textbf{SpiralMLP-B5} by adjusting the hyperparameters. In each model, 4 stages are integrated, and the spatial resolution is reduced while the channel dimension is increased along with the process. Thereby it facilitates effective down-sampling of spatial resolution and optimizes computational efficiency. A depiction of the PVT-style SpiralMLP architecture can be found in~\cref{fig:2} (a).

Furthermore, we have also developed variants modeled after the Swin architecture. The models are categorized into three types: \textbf{SpiralMLP-T (Tiny)}, \textbf{SpiralMLP-S (Small)}, and \textbf{SpiralMLP-B (Base)}. The structural details of both PVT-style and Swin-style will be further provided in the appendix. 


\section{Experiments}
\label{sec:experiment}

\begin{table}[tb]
    \label{tab:exp:cifar}
    \centering
    \resizebox{0.98\columnwidth}{!}{
    \begin{tabular}{@{}cccc@{}}
    \toprule
    Model & CIFAR-10(\%) & CIFAR-100(\%) & Params(M) \\
    \midrule
    \rowcolor{LightGray}
    \textbf{Spiral-B1 (ours)} & \textbf{95.6} & \textbf{78.6} & \textbf{14} \\
    CaiT~\cite{touvron2021going} & 94.9 & 76.9 & 9 \\
    MONet-T~\cite{cheng2024multilinear} & 94.8 & 77.2 & 10.3 \\
    Cycle-B1~\cite{chen2022cyclemlp} & 94.5 & 77.3 & 15 \\
    PiT~\cite{heo2021rethinking} & 94.2 &  75.0 & 7 \\
    Swin~\cite{liu2021swin} & 94.0 & 77.3 & 7 \\
    VGG19-bn~\cite{simonyan2015deep} & 94.0 & 72.2 & 39 \\
    ResNet50~\cite{he2015deep} & 93.7 & 77.4 & 24 \\
    ViT~\cite{dosovitskiy2021image} & 93.6 & 73.8 & 3 \\
    Swin-v2-T~\cite{liu2022swin} & 89.7 & 70.2 & 28 \\
    \bottomrule
    \end{tabular}}
    \caption{Top-1 accuracy achieved through training from scratch on both CIFAR-10 and CIFAR-100. }
\end{table}

\begin{table*}[tb]
    \label{tab:exp:1}
    \centering
    \resizebox{0.9\linewidth}{!}{
    \begin{tabular}{@{}cccc|cccc@{}}
    \hline
    \multirow{2}{*}{Model} & Top-1 & Params & FLOPs & \multirow{2}{*}{Model} & Top-1 & Params & FLOPs \\
    & Acc (\%) & (M) & (G) && Acc (\%) & (M) & (G)\\
    \hline
    \cellcolor{LightGray}\textbf{SpiralMLP-B5 (ours)} & \cellcolor{LightGray}\textbf{84.0} & \cellcolor{LightGray}\textbf{68} & \cellcolor{LightGray}\textbf{11.0} &\cellcolor{LightBlue} Swin-B~\cite{liu2021swin} &\cellcolor{LightBlue} 83.5 &\cellcolor{LightBlue} 88 &\cellcolor{LightBlue} 15.4 \\
    \cellcolor{LightGray}SpiralMLP-B4 (ours) & \cellcolor{LightGray}83.8 & \cellcolor{LightGray}46 & \cellcolor{LightGray}8.2&\cellcolor{LightBlue} gSwin-S~\cite{go2023gswin} &\cellcolor{LightBlue} 83.0 &\cellcolor{LightBlue} 19 &\cellcolor{LightBlue} 4.2 \\
    \cellcolor{LightGray}SpiralMLP-B (ours) & \cellcolor{LightGray}83.6 & \cellcolor{LightGray}67 & \cellcolor{LightGray}11.0 &\cellcolor{LightBlue} SimA-XCiT-S12/16~\cite{koohpayegani2024sima} &\cellcolor{LightBlue} 82.1 &\cellcolor{LightBlue} 26 &\cellcolor{LightBlue} 4.8 \\
    \cellcolor{LightGray}SpiraMLP-S (ours) & \cellcolor{LightGray}83.3 & \cellcolor{LightGray}56 & \cellcolor{LightGray}9.1 &\cellcolor{LightBlue} SimA-CvT-13~\cite{koohpayegani2024sima} &\cellcolor{LightBlue} 81.4 &\cellcolor{LightBlue} 20 &\cellcolor{LightBlue} 4.5 \\
    \cellcolor{LightRed}ATMNet-L~\cite{wei2022active} &\cellcolor{LightRed} 83.8 &\cellcolor{LightRed} 76 &\cellcolor{LightRed} 12.3 &\cellcolor{LightBlue} SimA-DeiT-S~\cite{koohpayegani2024sima} &\cellcolor{LightBlue} 79.8 &\cellcolor{LightBlue} 22 &\cellcolor{LightBlue} 4.6 \\
    \cellcolor{LightRed}HireMLP-Large~\cite{guo2021hiremlp} &\cellcolor{LightRed} 83.8 &\cellcolor{LightRed} 96 &\cellcolor{LightRed} 13.4 &\cellcolor{LightBlue} NOAH~\cite{li2024noah} &\cellcolor{LightBlue} 77.3 &\cellcolor{LightBlue} 26 &\cellcolor{LightBlue} - \\
    \cellcolor{LightRed}WaveMLP-B~\cite{tang2022image} &\cellcolor{LightRed} 83.6 &\cellcolor{LightRed} 63 &\cellcolor{LightRed} 10.2 &\cellcolor{LightBlue} CRATE-L~\cite{yu2024white} &\cellcolor{LightBlue} 71.3 &\cellcolor{LightBlue} 78 &\cellcolor{LightBlue} - \\
    \cellcolor{LightRed}MorphMLP-L~\cite{zhang2022morphmlp} &\cellcolor{LightRed} 83.4 &\cellcolor{LightRed} 76 &\cellcolor{LightRed} 12.5 &\cellcolor{LightBlue} CRATE-B~\cite{yu2024white} &\cellcolor{LightBlue} 70.8 &\cellcolor{LightBlue} 23 &\cellcolor{LightBlue} - \\
    \cline{5-8}
    \cellcolor{LightRed}MorphMLP-B~\cite{zhang2022morphmlp} &\cellcolor{LightRed} 83.2 &\cellcolor{LightRed} 58 &\cellcolor{LightRed} 10.2 & \cellcolor{LightYellow}DeepMAD-89M~\cite{shen2023deepmad} &\cellcolor{LightYellow} 84.0 &\cellcolor{LightYellow} 89 &\cellcolor{LightYellow} 15.4 \\
    \cellcolor{LightRed}CycleMLP-B~\cite{chen2022cyclemlp} &\cellcolor{LightRed} 83.4 &\cellcolor{LightRed} 88 &\cellcolor{LightRed} 15.2 &\cellcolor{LightYellow} DeepMAD-50M~\cite{shen2023deepmad} &\cellcolor{LightYellow} 83.9 &\cellcolor{LightYellow} 50 &\cellcolor{LightYellow} 8.7 \\
    \cellcolor{LightRed}CycleMLP-B5~\cite{chen2022cyclemlp} &\cellcolor{LightRed} 83.1 &\cellcolor{LightRed} 76 &\cellcolor{LightRed} 12.3 &\cellcolor{LightYellow}  EfficientNet-B4~\cite{tan2020efficientnet} &\cellcolor{LightYellow} 82.6 &\cellcolor{LightYellow} 19 &\cellcolor{LightYellow} 4.2 \\
    \cellcolor{LightRed}sMLP-B~\cite{tang2022sparse} &\cellcolor{LightRed} 83.4 &\cellcolor{LightRed} 66 &\cellcolor{LightRed} 14.0 &\cellcolor{LightYellow} VanillaNet-13-1.5~\cite{chen2024vanillanet} &\cellcolor{LightYellow} 82.5 &\cellcolor{LightYellow} 128 &\cellcolor{LightYellow} 26.5 \\
    \cellcolor{LightRed}ASMLP-B~\cite{lian2022asmlp} &\cellcolor{LightRed} 83.3 &\cellcolor{LightRed} 88 &\cellcolor{LightRed} 15.2  &\cellcolor{LightYellow} VanillaNet-13~\cite{chen2024vanillanet} &\cellcolor{LightYellow} 82.1 &\cellcolor{LightYellow} 59 &\cellcolor{LightYellow} 11.9 \\
    \cellcolor{LightRed}gMLP~\cite{liu2021pay} &\cellcolor{LightRed} 81.6 &\cellcolor{LightRed} 45 &\cellcolor{LightRed} 31.6 &\cellcolor{LightYellow} HGRN-DeiT-Small~\cite{qin2024hierarchically} &\cellcolor{LightYellow} 80.1 &\cellcolor{LightYellow} 24 &\cellcolor{LightYellow} - \\
    \cellcolor{LightRed}ConvMixer-1536/20~\cite{trockman2022patches} &\cellcolor{LightRed} 81.4 &\cellcolor{LightRed} 52&\cellcolor{LightRed} - &\cellcolor{LightYellow} {HGRN-DeiT-Tiny}~\cite{qin2024hierarchically} &\cellcolor{LightYellow} 74.4 &\cellcolor{LightYellow} 6 &\cellcolor{LightYellow} - \\
    \cellcolor{LightRed}ConvMixer-1536/20~\cite{trockman2022patches} & \cellcolor{LightRed}80.4 & \cellcolor{LightRed}49 & \cellcolor{LightRed}- &\cellcolor{LightYellow} ResNet-50~\cite{he2015deep} &\cellcolor{LightYellow} 75.3 &\cellcolor{LightYellow} 25 &\cellcolor{LightYellow} 3.8 \\
    \cline{5-8}
    \cellcolor{LightRed}MONet-S~\cite{cheng2024multilinear} & \cellcolor{LightRed}81.3 & \cellcolor{LightRed}33 & \cellcolor{LightRed}6.8 &\cellcolor{LightPurple} Vim-S~\cite{zhu2024vision} &\cellcolor{LightPurple} 80.5 &\cellcolor{LightPurple} 26 &\cellcolor{LightPurple} - \\
    \cellcolor{LightRed}MONet-T~\cite{cheng2024multilinear} & \cellcolor{LightRed}77.0 & \cellcolor{LightRed}10 & \cellcolor{LightRed}2.8 &\cellcolor{LightPurple} {Vim-Ti}~\cite{zhu2024vision} &\cellcolor{LightPurple} 78.3 &\cellcolor{LightPurple} 7 &\cellcolor{LightPurple} - \\
    \cellcolor{LightRed}ResMLP-B24~\cite{touvron2021resmlp} & \cellcolor{LightRed}81.0 & \cellcolor{LightRed}116 & \cellcolor{LightRed}23.0 &\cellcolor{LightPurple} M2-ViT-b~\cite{fu2024monarch} &\cellcolor{LightPurple} 79.5 &\cellcolor{LightPurple} 45 &\cellcolor{LightPurple} - \\
    \cellcolor{LightRed}S$^2$MLP-deep~\cite{yu2021s2mlp} & \cellcolor{LightRed}80.7 & \cellcolor{LightRed}51 & \cellcolor{LightRed}10.5 & \cellcolor{LightPurple}ViT-b-Monarch~\cite{fu2024monarch} & \cellcolor{LightPurple}78.9 & \cellcolor{LightPurple}33 & \cellcolor{LightPurple}- \\
    \cellcolor{LightRed}S$^2$MLP-wide~\cite{yu2021s2mlp} & \cellcolor{LightRed}80.0 & \cellcolor{LightRed}71 & \cellcolor{LightRed}14.0 & \cellcolor{LightPurple}HyenaViT-b~\cite{poli2023hyena} & \cellcolor{LightPurple}78.5 & \cellcolor{LightPurple}88 & \cellcolor{LightPurple}- \\
    \cline{5-8}
    \cellcolor{LightRed}ConvMLP-L~\cite{li2023convmlp} & \cellcolor{LightRed}80.2 & \cellcolor{LightRed}43 & \cellcolor{LightRed}9.9 &\cellcolor{LightRed}RepMLP-Res50-g8/8~\cite{ding2021repmlp} & \cellcolor{LightRed}76.4 & \cellcolor{LightRed}59 & \cellcolor{LightRed}12.7 \\
    \cellcolor{LightRed}ConvMLP-M~\cite{li2023convmlp} & \cellcolor{LightRed}79.0 & \cellcolor{LightRed}17 & \cellcolor{LightRed}4.0 &\cellcolor{LightRed}MLPMixer-B/16~\cite{tolstikhin2021mlpmixer} & \cellcolor{LightRed}76.4 & \cellcolor{LightRed}59 & \cellcolor{LightRed}12.7 \\
    \cellcolor{LightRed}RepMLP-Res50-g4/8~\cite{ding2021repmlp} & \cellcolor{LightRed}80.1 & \cellcolor{LightRed}87 & \cellcolor{LightRed}8.2 &\cellcolor{LightRed}AFFNet~\cite{huang2023adaptive} & \cellcolor{LightRed}79.8 & \cellcolor{LightRed}6 & \cellcolor{LightRed}1.5 \\
    
    \hline
    \end{tabular}}
    \caption{Top-1 accuracy on ImageNet-1k, with $224\times224$ as the input resolution. In terms of background colors, \colorbox{LightRed}{\phantom{}}, \colorbox{LightBlue}{\phantom{}}, \colorbox{LightYellow}{\phantom{}}, \colorbox{LightPurple}{\phantom{}} denote MLPs, Transformers, CNNs and State-Space Models, respectively. }
\end{table*}

\begin{figure*}[tb]
  \centering
  \fbox{\includegraphics[width=0.95\linewidth]{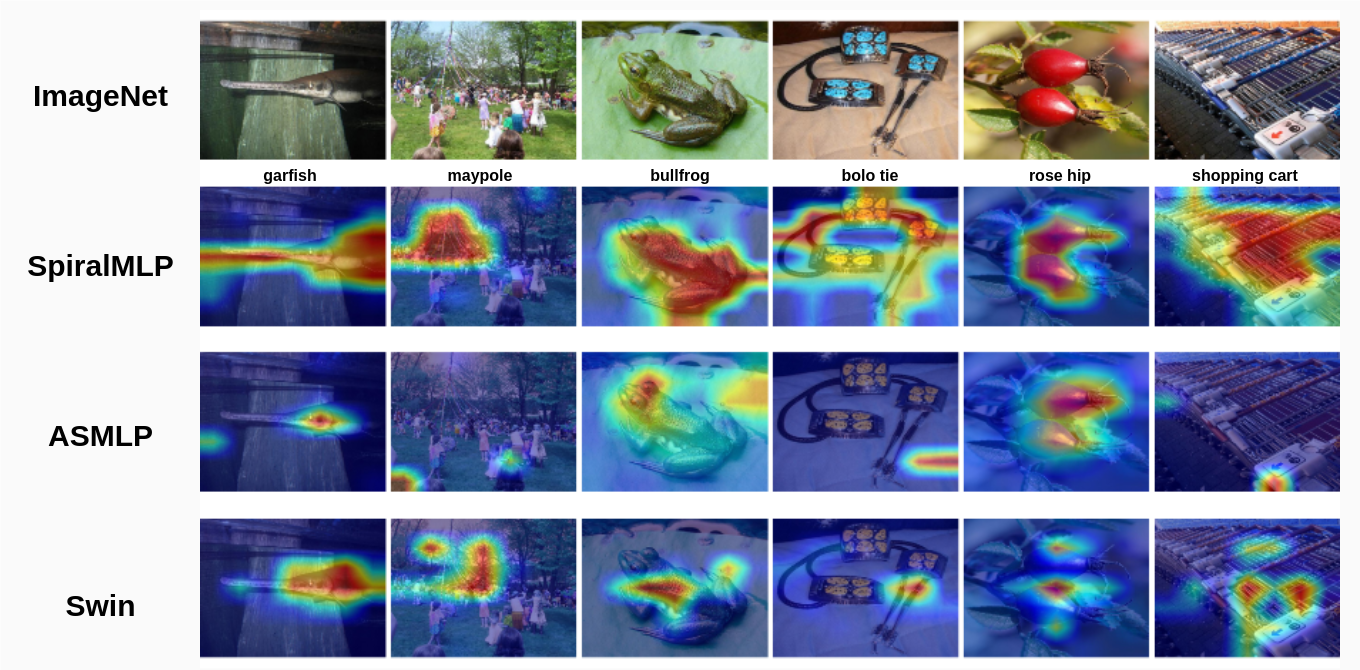}}
   \caption{In contrast to ASMLP~\cite{lian2022asmlp} and Swin~\cite{liu2021swin}, our SpiralMLP demonstrates superior object-focused attention. SpiralMLP exhibits enhanced sensitivity, especially for elongated or curved objects. The backbones employed for heatmaps generation are SpiralMLP-B5, ASMLP-B and Swin-B. The images are sourced from the ImageNet-1k~\cite{russakovsky2015imagenet} validation dataset, with corresponding labels.}
   \label{fig:4}
\end{figure*}

We initially perform experiments with SpiralMLP-B1 on CIFAR-10~\cite{Krizhevsky09learningmultiple} and CIFAR-100~\cite{Krizhevsky09learningmultiple}, comparing it against architectures of similar scale, including MLPs, CNNs, and Transformers. The outcomes are presented in~\cref{tab:exp:cifar}, all of the models are trained from scratch. 

We extend our experimentation to include image classification on ImageNet-1k~\cite{russakovsky2015imagenet}, as well as object detection and instance segmentation on the COCO~\cite{lin2015microsoft}. Furthermore, we assess its semantic segmentation capabilities on ADE20K~\cite{zhou2018semantic}.

\subsection{Image Classification on ImageNet-1k}
\subsubsection{Settings}
Our implementation primarily draws from DeiT~\cite{touvron2021training}. The training is 4 NVIDIA A100 GPUs for a total of 300 epochs. The overall batch size is 512 and we employ the Top-1 accuracy for image classification.

\subsubsection{Comparison with MLPs}
As shown in~\cref{tab:exp:1} \colorbox{LightRed}{\phantom{}}, SpiralMLP-B achieves a Top-1 accuracy of $84.0\%$ on the ImageNet-1k, with the input resolution of $224\times224$. This performance notably exceeds that of the best-performing models of ATMNet-L~\cite{wei2022active}, HireMLP-Large~\cite{guo2021hiremlp}, WaveMLP-B~\cite{tang2022image}, MorphMLP-L~\cite{zhang2022morphmlp} and CycleMLP-B~\cite{chen2022cyclemlp}, by +$0.2\%$, +$0.2\%$, +$0.4\%$, +$0.6\%$ and +$0.6\%$, respectively. Furthermore, compared to the S$^2$MLP-wide~\cite{yu2021s2mlp}, which has a similar model size with $71$M parameters, SpiralMLP surpasses it by +$4.0\%$ with only $68$M parameters. In addition to the advantage on the model size, SpiralMLP also demonstrates potential balance between computational efficiency and accuracy. It is evident that among a cohort of models with accuracy exceeding $83\%$ (including ATMNet-L~\cite{wei2022active}, HireMLP-Large~\cite{guo2021hiremlp}, WaveMLP-B~\cite{tang2022image}, MorphMLP-B~\cite{zhang2022morphmlp}, MorphMLP-L~\cite{zhang2022morphmlp}, CycleMLP-B~\cite{chen2022cyclemlp}, CycleMLP-B5~\cite{chen2022cyclemlp}, sMLP-B~\cite{tang2022sparse} and ASMLP-B~\cite{lian2022asmlp}, SpiralMLP-B5 stands out due to a lower FLOPs of $11.0$G and the highest accuracy.

\begin{table*}[tb]
    \label{tab:exp:2}
    \centering
    \resizebox{0.85\linewidth}{!}{
    \begin{tabular}{@{}c|cc|ccc|ccc@{}}
    \hline
    \multirow{3}{*}{\vspace{1em}BackBone} & \multicolumn{8}{c}{RetinaNet $1\times$}\\\cline{2-9}
    & Params (M) & FLOPs (G) & AP & AP$_{50}$ & AP$_{75}$ & AP$_S$ & AP$_M$ & AP$_L$\\
    \hline
    
    ResNet101~\cite{he2015deep} & 56.7 & 492.2 & 38.5 & 57.8 & 41.2 & 21.4 & 42.6 & 51.1 \\
    ConvMLP-L~\cite{li2023convmlp} & 52.9 & - & 40.2 & 59.3 & 43.3 & 23.5 & 43.8 & 53.3 \\
    ResNeXt101-64x4d~\cite{xie2017aggregated} & 95.5 &-& 41.0 & 60.9 & 44.0 & 23.9 & 45.2 & 54.0 \\
    CycleMLP-B5~\cite{chen2022cyclemlp} & 85.9 & 360.3 & 42.7 & 63.3 & 45.3 & 24.1 & 46.3 & 57.4 \\
    ATMNet-L~\cite{wei2022active} & 86.0 & 405.0 & 46.1 & 67.4 & 49.4 & 29.9 & 50.1 & 61.0 \\
    PVTv2-B5~\cite{wang2022pvt} & 91.7 &-& 46.2 & 67.1 & 49.5 & 28.5 & 50.0 & 62.5 \\
    \rowcolor{LightGray}
    \textbf{SpiralMLP-B5 (ours)} & \textbf{79.8} & \textbf{325.0} & \textbf{46.5} & \textbf{67.7} & \textbf{49.8} & \textbf{30.3} & \textbf{50.8} & \textbf{62.8} \\
    \hline
    \multirow{3}{*}{\vspace{1em}BackBone} & \multicolumn{8}{c}{Mask R-CNN $1\times$}\\\cline{2-9}
    & Params (M) & FLOPs (G) & AP & AP$_{50}$ & AP$_{75}$ & AP$_S$ & AP$_M$ & AP$_L$\\
    \hline
    ResNet101~\cite{he2015deep} & 63.2 & - & 40.4 & 61.1 & 44.2 & 36.4 & 57.7 & 38.8 \\
    ConvMLP-L~\cite{li2023convmlp} & 62.2 & - & 41.7 & 62.8 & 45.5 & 38.2 & 59.9 & 41.1 \\
    Swin-T~\cite{liu2021swin} & 47.8 & 267.0 & 42.7 & 65.2 & 46.8 & 39.3 & 62.2 & 42.2 \\
    ResNeXt101-64x4d~\cite{xie2017aggregated} & 101.9 &-& 42.8 & 63.8 & 47.3 & 38.4 & 60.6 & 41.3 \\
    VanillaNet-13~\cite{chen2024vanillanet} & 76.3 & 421.0 & 42.9 & 65.5 & 46.9 & 39.6 & 62.5 & 42.2 \\ 
    CycleMLP-B5~\cite{chen2022cyclemlp} & 95.3 &-& 44.1 & 65.5 & 48.4 & 40.1 & 62.8 & 43.0 \\
    HireMLP-Large (1x)~\cite{guo2021hiremlp} & 155.2 & 443.5 & 45.9 & 67.2 & 50.4 & 41.7 & 64.7 & 45.3 \\
    PVTv2-B5~\cite{wang2022pvt} & 101.6 & 334.5 & 47.4 & 68.6 & 51.9 & 42.5 & 65.7 & 46.0 \\
    ATMNet-L~\cite{wei2022active} & 96.0 & 424.0 & 47.4 & 69.9 & 52.0 & 43.2 & 67.3 & 46.5 \\
    \rowcolor{LightGray}
    \textbf{SpiralMLP-B (ours)} & \textbf{89.1} & \textbf{342.0} & \textbf{47.8} & \textbf{71.6} & \textbf{53.2} & \textbf{43.6} & \textbf{69.3} & \textbf{47.2} \\
    \hline
    \end{tabular}}
    \caption{Object detection performance with RetinaNet $1\times$ and MASK R-CNN $1\times$ on the COCO validation dataset, all of the backbones are pretrained on the ImageNet-1k. The FLOPs are evaluated at a resolution of $1280\times800$. The entries are sorted in ascending order based on AP performance.}
\end{table*}

\subsubsection{Comparison with other SOTAs}
SpiralMLP remains competitive over Transformers, CNNs and State-Space Models, particularly in significantly reducing the number of parameters and the FLOPs as referenced in~\cref{tab:exp:1} \colorbox{LightBlue}{\phantom{}}, \colorbox{LightYellow}{\phantom{}}, \colorbox{LightPurple}{\phantom{}}. For instance, when comparing SpiralMLP-B5 to CNNs \colorbox{LightYellow}{\phantom{}}, it outperforms VanillaNet-13-1.5~\cite{chen2024vanillanet} by +$1.5\%$ and has the same performance to DeepMAD-89M~\cite{shen2023deepmad}. When comparing between State-Space Models \colorbox{LightPurple}{\phantom{}} and SpiralMLP-B4 as well as SpiralMLP-S, SpiralMLP demonstrates a notable performance improvement of approximate +$4.0\%$. Furthermore, when comparing with the Transformers \colorbox{LightBlue}{\phantom{}}, SpiralMLP-B5 has nearly $20$M fewer parameters than Swin-B~\cite{liu2021swin} while achieving +$0.5\%$ higher in accuracy. Particularly the vision transformers continue struggling with quadratic complexity. And in order to better demonstrate, we visualize the heatmaps in~\cref{fig:4} in comparison with the performance of ASMLP~\cite{lian2022asmlp} and Swin~\cite{liu2021swin}.

\subsection{Object Detection and Instance Segmentation on COCO}
\subsubsection{Settings}
We conduct object detection and instance segmentation experiments on COCO~\cite{lin2015microsoft}, wherein we demonstrate SpiralMLP with PVT and Swin architectures, adopting two distinct configurations. We leverage SpiralMLP-B5 and Spiral-B with the pretrained weights on ImageNet-1k~\cite{russakovsky2015imagenet} as the backbones, together with Xavier initialization~\cite{pmlr-v9-glorot10a} applied to the newly added layers.

\subsubsection{Results}
\begin{figure}[tb]
  \centering
  \fbox{\includegraphics[width=0.95\linewidth]{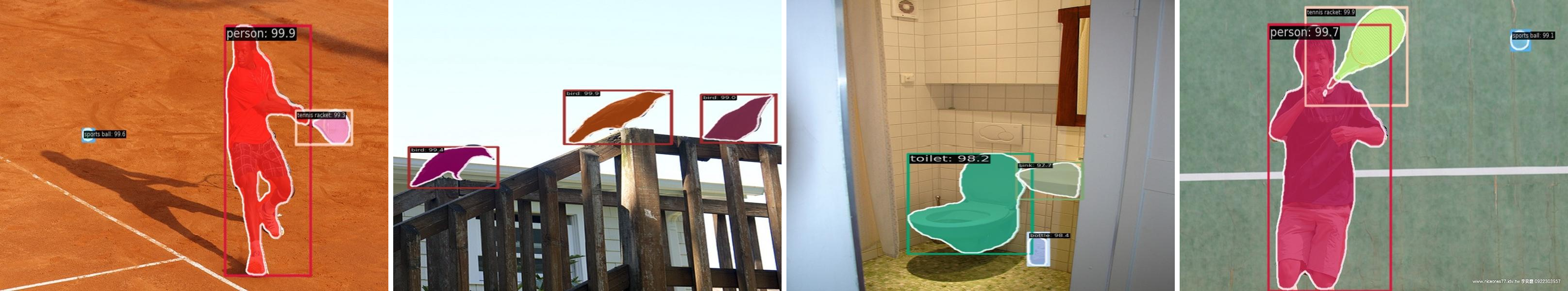}}
   \caption{Several examples of the object detection and instance segmentation from COCO~\cite{lin2015microsoft} test dataset.}
   \label{fig:5}
\end{figure}

Comparative results are detailed in~\cref{tab:exp:2}, where we employ either RetinaNet~\cite{lin2018focal} or Mask R-CNN~\cite{he2018mask} as the detection framework. When comparing under the RetinaNet $1\times$, SpiralMLP-B5 stands out in terms of the highest AP. In particular, it achieves +$0.3\%$ higher than PVTv2-B5, with -$11.9$M fewer parameters. In the context of Mask R-CNN $1\times$, SpiralMLP-B outperforms ATMNet-L by +$0.4\%$ in AP, alongside a reduction of $6.9$M in model parameters. Visual representations of the object detection and instance segmentation are presented in~\cref{fig:5}.

\begin{figure}[tb]
  \centering
  \fbox{\includegraphics[width=0.95\linewidth]{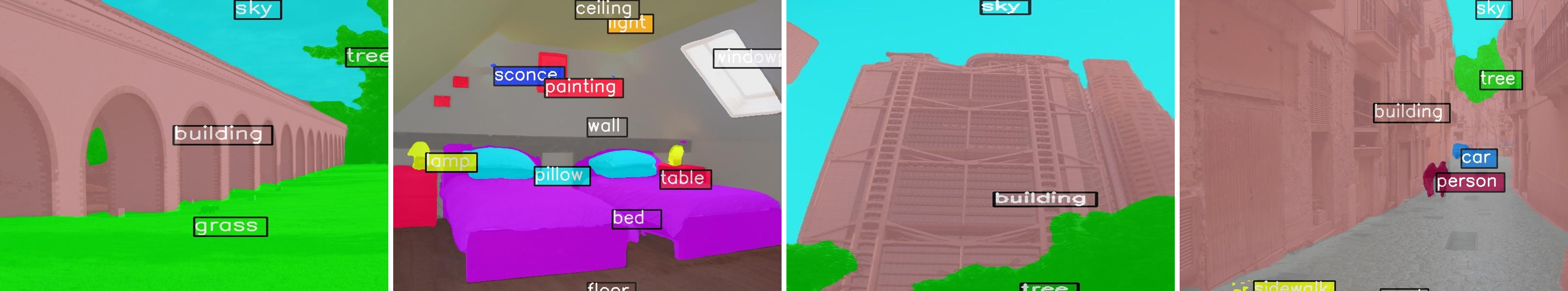}}
   \caption{Several examples of the semantic segmentation from ADE20K~\cite{zhou2018semantic} validation dataset.}
   \label{fig:6}
\end{figure}

\begin{table*}[tb]
    \label{tab:exp:semseg}
    \centering
    \resizebox{0.85\linewidth}{!}{
    \begin{tabular}{@{}cccc|cccc@{}}
    \hline
         \multirow{2}{*}{Model} & \multicolumn{3}{c|}{Semantic FPN} & \multirow{2}{*}{Model} & \multicolumn{3}{c}{UperNet}\\
          & Params & FLOPs & mIoU & & Params & FLOPs & mIoU\\
    \hline
        ResNet101~\cite{he2015deep} & 47.5 & 10.1 & 38.8 & DeepMAD-29M*~\cite{shen2023deepmad} & 27 & 56 & 46.9 \\
        ConvMLP-L~\cite{li2023convmlp} & 46.3 & - & 40.0 & HireMLP-Large~\cite{guo2021hiremlp} & 127 & 1125 & 48.8 \\
        ResNeXt101-64x4d~\cite{xie2017aggregated} & 86.4 & 103.9 & 40.2 & Focal-B~\cite{yang2021focal} & 126 & - & 49.0\\
        CycleMLP-B5~\cite{chen2022cyclemlp} & 79.4 & 86.0 & 45.5 &ConvNeXt-T~\cite{liu2022convnet}&82&-&48.7\\
        MorphMLP-B~\cite{zhang2022morphmlp} & 59.3 & 76.8 & 45.9 & ConvNeXt-B~\cite{liu2022convnet} & 122 & - & 49.1 \\
        Swin-B~\cite{liu2021swin} & 91.2 & 107.0 & 46.0 & AS-MLP-B~\cite{lian2022asmlp} & 121 & 1166 & 49.5 \\
        Twins-L~\cite{chu2021twins} & 103.7 & 102.0 & 46.7 & Swin-B~\cite{liu2021swin} & 121 & 1188 & 49.7 \\
        ConvNeXt-T~\cite{liu2022convnet} & 27.8 & 93.2 & 46.7 & CycleMLP-B~\cite{chen2022cyclemlp} & 121 & 1166 & 49.7 \\
        ATMNet-L~\cite{wei2022active} & 79.8 & 86.6 & 48.1 & ATMNet-L~\cite{wei2022active} & 108 & 1106 & 50.1 \\
        PVTv2-B5~\cite{wang2022pvt} & 85.7 & 91.1 & 48.7 & FocalNet-B(LRF)~\cite{yang2022focal} & 126 & - & 50.5 \\
        \cellcolor{LightGray}\textbf{SpiralMLP-B5 (ours)} & \cellcolor{LightGray}\textbf{73.2} & \cellcolor{LightGray}\textbf{75.5} & \cellcolor{LightGray}\textbf{48.9} & \cellcolor{LightGray}\textbf{SpiralMLP-B (ours)} & \cellcolor{LightGray}\textbf{100} & \cellcolor{LightGray}\textbf{1061} & \cellcolor{LightGray}\textbf{50.7} \\
    \hline
    \end{tabular}}
    \caption{Semantic segmentation performance on ADE20K validation dataset with Semantic FPN as well as UperNet. When evaluated with Semantic FPN, the FLOPs are measured at a resolution of $512\times512$. When evaluated with UperNet, the FLOPs are measured at a resolution of $2048\times512$. The entries are sorted in ascending order based on mIOU performance.}
\end{table*}

\subsection{Semantic Segmentation on ADE20K}
\subsubsection{Settings}
We perform semantic segmentation on the ADE20K dataset using UperNet~\cite{xiao2018unified} and Semantic FPN~\cite{kirillov2019panoptic} as the frameworks. For the backbones, we employ SpiralMLP-B5 and SpiralMLP-B, with the weights pretrained on ImageNet-1k. Additionally, the newly add layers are initialized with Xavier~\cite{pmlr-v9-glorot10a}.

\subsubsection{Results}
As depicted in~\cref{tab:exp:semseg}, SpiralMLP still exhibits comparable performance when integrated with Semantic FPN and UperNet for semantic segmentation tasks. In the Semantic FPN evaluations, SpiralMLP-B5 surpasses its closest competitor, PVTv2-B5, by +$0.2\%$, and exceeds the second-best model, ATMNet-L, by +$0.6\%$. When integrated with UperNet, SpiralMLP-B still emerges as the top-performing model, outperforming  FocalNet-B(LRF)~\cite{yang2022focal} by +$0.2\%$ and ATMNet-L~\cite{wei2022active} by +$0.6\%$. Visual representations of the semantic segmentation are presented in~\cref{fig:6}. 


\section{Ablation}
\begin{figure}[t]
    \centering
    \fbox{\includegraphics[width=0.9\columnwidth]{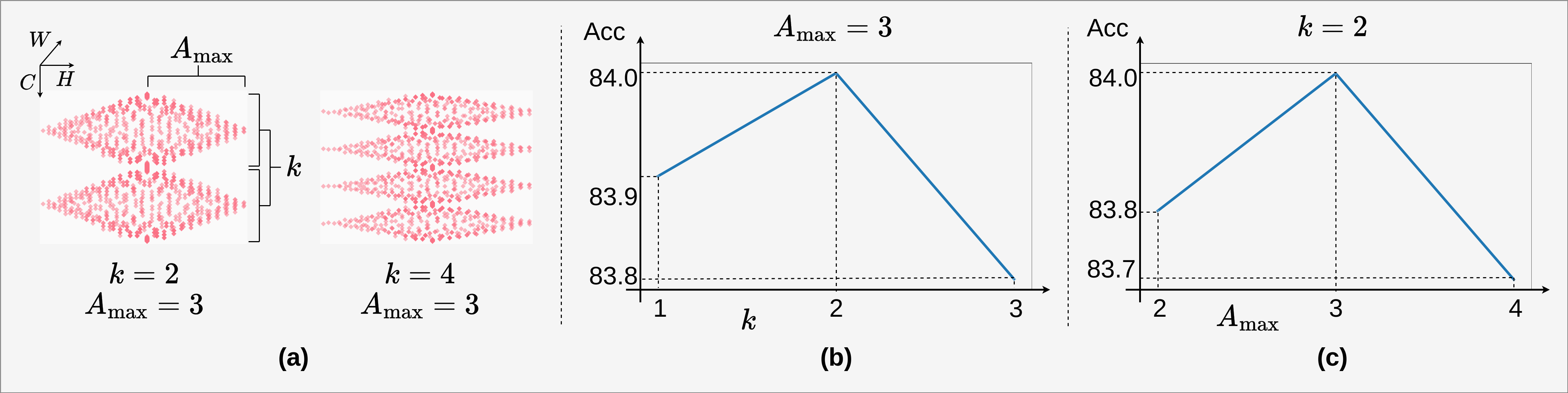}}
    \caption{Visualization of varying $k$ on spiral trajectory as described by~\cref{eq:abn:3,eq:abn:4}, while maintaining a constant $A_{\text{max}}=3$. }
    \label{fig:abn:1}
\end{figure}

\begin{table}
    \centering
    \begin{tabular}{c|c|c|c|c|c}
    \hline
    \textbf{Case 1} & \multicolumn{5}{c}{$k$} \\
    \hline
    $A_\text{max}=3$ & 1 & 2 & 3 & 4 & 5 \\
    \hline
    Acc(\%) & 83.9 & 84.0 & 83.8 & 83.6 & 83.3 \\    
    \hline
    \textbf{Case 2} & \multicolumn{5}{c}{$A_\text{max}$} \\
    \hline
    $k=2$ & 2 & 3 & 4 & 5 & 6 \\
    \hline
    Acc(\%) & 83.8 & 84.0 & 83.7 & 83.4 & 82.9 \\    
    \hline
    \end{tabular}
    \caption{Experiments on $k$ and $A_\text{max}$. After reaching their respective peaks, both trends show a rapid decline.} 
    \label{tab:kamax}
\end{table}

\subsection{Update the Offset Functions}
\label{sec:abn:update}

The offset functions $\phi_i(\cdot)$ and $\phi_j(\cdot)$ (\cref{eq:6,eq:7}) are originally designed into a two-partition pattern, and we further expand them to a more generic multi-partition pattern. To incorporate this update, we introduce $k$ as the number of partitions along the channel dimension. The partitions can be defined as follows:
\begin{align}
    P=\left\{0,\frac{C_{\text{in}}}{k},\frac{2*C_{\text{in}}}{k},\dots,C_{\text{in}}\right\}\label{eq:abn:1}
\end{align}

By introducing $k$ and considering individual partition, we can create multiple spiral structures that capture the characteristics of each partition along the channel dimension. Furthermore, we define the length of the partition as $C_w=\frac{C_{\text{in}}}{k}$, which is the distance between two adjacent endpoints, then the amplitude function~\cref{eq:8} is updated to:

\begin{align}
A^*(c)=\begin{cases}
\left\lfloor\frac{2A_{\text{max}}}{C_w}(c-iC_w)\right\rfloor, &0\le c<\frac{iC_w}{2} \\
\left\lfloor(2A_{\text{max}}-\frac{2A_{\text{max}}}{C_w})(c-iC_w)\right\rfloor, &\frac{iC_w}{2}\le c\le iC_w
\end{cases}
\label{eq:abn:2}
\end{align}

where, $i\in[0, 1\dots,k-1]$ represents the $i^{\text{th}}$ partition in partitions $P$ (\cref{eq:abn:1}), and $c$ in~\cref{eq:8} is replaced by $z$ within the $i^{\text{th}}$ partition. Accordingly,~\cref{eq:6,eq:7} are updated as:
\begin{align}
\phi^*_i(c)&=A^*(c)\cos(\frac{c*2\pi}{T})\label{eq:abn:3} \\
\phi^*_j(c)&=A^*(c)\sin(\frac{c*2\pi}{T})\label{eq:abn:4}
\end{align}

We also provide the visualizations of~\cref{eq:abn:3,eq:abn:4}, as depicted in~\cref{fig:abn:1}, showcasing variations with different numbers of partitions $k$.

\begin{table*}[th]
    \label{tab:abn:fc}
    \centering
    \begin{tabular}{c|c|c|c|c|c}
    \hline
     & \cellcolor{LightGray}SpiralFC & \multirow{2}{*}{PATM} & \multirow{2}{*}{ATMLayer} & \multirow{2}{*}{CycleFC} & \multirow{2}{*}{RandomFC} \\
     & \cellcolor{LightGray}\textbf{(ours)} &&&\\ 
    \hline
    Acc (\%) & \cellcolor{LightGray}\textbf{95.6} & 95.3 & 95.2 & 94.7 & 94.5 \\
    Params (M) & \cellcolor{LightGray}\textbf{14} & 17 & 15 & 15 & 14 \\
    \hline
    \end{tabular}
    \caption{The accuracy on CIFAR-10, each Fully-Connected Layer is configured into the SpiralMLP-B1 architecture and is trained from scratch. } 
\end{table*}

\subsection{Ablation Study on $k$}
We updated the offset functions $\phi^*_i(\cdot)$ and $\phi_j^*(\cdot)$ (\cref{eq:abn:3,eq:abn:4}) to analyze how varying the number of partitions $k$ affects Top-1 Accuracy on ImageNet-1k. Results, shown in~\cref{tab:kamax} with a constant maximum amplitude $A_{\text{max}}$ of $3$, indicate that accuracy initially rises, peaks at $k=2$, then decreases.

This trend suggests that different $k$ values alter the focus on the peripheral regions of the receptive field, where $k=2$ results in denser clustering of feature points along the edges compared to $k=4$, as seen in~\cref{fig:abn:1}. Lower $k$ values cause a dense, narrow concentration of features, while higher values disperse them too widely, potentially reducing model effectiveness.

\subsection{Ablation Study on $A_{\text{max}}$}
We investigate several cases when the maximum amplitude $A_{\text{max}}$ takes various values. From the results shown in~\cref{tab:kamax}, we observe an initial improvement in the Top-1 Accuracy on ImageNet-1k. However, a decline becomes evident once the $A_{\text{max}}$ exceeds $3$. 

Similarly, the underlying reason is that, as $A_{\text{max}}$ increases, the receptive field's extent expands. However, due to the characteristics of Spiral FC, the number of selected feature points remains constant at $C_{\text{in}}$. Consequently, a larger $A_{\text{max}}$ results in a more sparse distribution of feature points. If $A_{\text{max}}$ is too small, the Spiral FC may fail to encompass a sufficient number of neighboring features. On the other hand, if $A_{\text{max}}$ is excessively large, the Spiral FC might not effectively capture detailed information within the receptive field.

Although the discrete experimental design does not guarantee the discovery of optimal hyperparameters, it indeed facilitates the insight of underlying trends and tendencies.

\subsection{Ablation Study on Fully-Connected Layers}
To illustrate the effectiveness of Spiral FC, we perform experiments on the CIFAR-10~\cite{Krizhevsky09learningmultiple} using SpiralMLP-B1\footnote{The configuration of SpiralMLP-B1 is demonstrated in Appendix. } as the base architecture. In these experiments, the Spiral FC is substituted with various alternatives, including \texttt{PATM} from WaveMLP~\cite{tang2022image}, \texttt{ATMLayer} from ATM~\cite{wei2022active}, \texttt{CycleFC} from CycleMLP~\cite{chen2022cyclemlp} and a \texttt{RandomFC}. The Random FC is architecturally identical to Spiral FC, except that the offset function is generated randomly. 

\subsection{Latency Analysis}

To highlight the speed efficiency of Spiral FC, we assess its performance across various input resolutions compared to other proposed architectures. We adopt the format from EfficientFormer~\cite{li2022efficientformer} and detail the latency analysis in ~\cref{tab:abn:latency}. We present SpiralMLP-B4 and SpiralMLP-B5 with several other architectures closely related to our study and specifically at the $224^2$ resolution. For a comprehensive latency comparison across different scenarios, please refer to the Appendix.

\begin{table}[t]
    \label{tab:abn:latency}
    \centering
    \resizebox{0.9\columnwidth}{!}{
    \begin{tabular}{cccc}
    \toprule
    Model & Params(M)  & Latency(ms) \\
    \midrule
    \cellcolor{LightGray}SpiralMLP-B4 & \cellcolor{LightGray}46  & \cellcolor{LightGray}47.00 \\
    \cellcolor{LightGray}SpiralMLP-B5 & \cellcolor{LightGray}68  & \cellcolor{LightGray} 39.22 \\
    \midrule
    CycleMLP-B4~\cite{chen2022cyclemlp} & 52  & 57.94 \\
    CycleMLP-B5 & 76  & 48.38\\
    \midrule 
    ATM-B~\cite{wei2022active} & 52 & 64.77  \\
    ATM-L & 76 & 54.09 \\
    \midrule
    PVTv2-B4~\cite{wang2022pvt} & 63 & 43.96 \\ 
    PVTv2-B5 & 82 & 55.71 \\
    \bottomrule
    \end{tabular}}
    \caption{Inference latency measured in \textit{milliseconds} on one A100. SpiralMLP outperforms other models of similar size in terms of speed. A single image of with $224^2$ resolution serves as the input.} 
\end{table}


\section{Conclusion and Future Work}
In this paper, we present Spiral FC, part of Spiral Mixing designed to replace traditional Token Mixing. We introduce SpiralMLP, a new computer vision framework compatible with PVT-style and Swin-style architectures. SpiralMLP performs comparably to leading models while using fewer parameters and less computational power.

We believe we are the first to use carefully designed offset functions to capture comprehensive feature information, setting us apart from models like CyelcMLP~\cite{chen2022cyclemlp}, ASMLP~\cite{lian2022asmlp} and ATM~\cite{wei2022active}, which focus on optimizing cross-like layers. Given its strong performance, further research into optimizing SpiralMLP's hyperparameters could lead to even more efficient information capture.

{\small
\bibliographystyle{ieee_fullname}
\bibliography{egbib}

\begin{thebibliography}{10}\itemsep=-1pt

\bibitem{ba2016layer}
Jimmy~Lei Ba, Jamie~Ryan Kiros, and Geoffrey~E. Hinton.
\newblock Layer normalization, 2016.

\bibitem{bao2022beit}
Hangbo Bao, Li Dong, Songhao Piao, and Furu Wei.
\newblock Beit: Bert pre-training of image transformers, 2022.

\bibitem{brown2020language}
Tom~B. Brown, Benjamin Mann, Nick Ryder, Melanie Subbiah, Jared Kaplan, Prafulla Dhariwal, Arvind Neelakantan, Pranav Shyam, Girish Sastry, Amanda Askell, Sandhini Agarwal, Ariel Herbert-Voss, Gretchen Krueger, Tom Henighan, Rewon Child, Aditya Ramesh, Daniel~M. Ziegler, Jeffrey Wu, Clemens Winter, Christopher Hesse, Mark Chen, Eric Sigler, Mateusz Litwin, Scott Gray, Benjamin Chess, Jack Clark, Christopher Berner, Sam McCandlish, Alec Radford, Ilya Sutskever, and Dario Amodei.
\newblock Language models are few-shot learners, 2020.

\bibitem{chen2024vanillanet}
Hanting Chen, Yunhe Wang, Jianyuan Guo, and Dacheng Tao.
\newblock Vanillanet: the power of minimalism in deep learning.
\newblock {\em Advances in Neural Information Processing Systems}, 36, 2024.

\bibitem{mmdetection}
Kai Chen, Jiaqi Wang, Jiangmiao Pang, Yuhang Cao, Yu Xiong, Xiaoxiao Li, Shuyang Sun, Wansen Feng, Ziwei Liu, Jiarui Xu, Zheng Zhang, Dazhi Cheng, Chenchen Zhu, Tianheng Cheng, Qijie Zhao, Buyu Li, Xin Lu, Rui Zhu, Yue Wu, Jifeng Dai, Jingdong Wang, Jianping Shi, Wanli Ouyang, Chen~Change Loy, and Dahua Lin.
\newblock {MMDetection}: Open mmlab detection toolbox and benchmark.
\newblock {\em arXiv preprint arXiv:1906.07155}, 2019.

\bibitem{chen2022cyclemlp}
Shoufa Chen, Enze Xie, Chongjian Ge, Runjian Chen, Ding Liang, and Ping Luo.
\newblock Cyclemlp: A mlp-like architecture for dense prediction, 2022.

\bibitem{cheng2024multilinear}
Yixin Cheng, Grigorios~G Chrysos, Markos Georgopoulos, and Volkan Cevher.
\newblock Multilinear operator networks.
\newblock {\em arXiv preprint arXiv:2401.17992}, 2024.

\bibitem{chu2021twins}
Xiangxiang Chu, Zhi Tian, Yuqing Wang, Bo Zhang, Haibing Ren, Xiaolin Wei, Huaxia Xia, and Chunhua Shen.
\newblock Twins: Revisiting the design of spatial attention in vision transformers.
\newblock {\em Advances in Neural Information Processing Systems}, 34:9355--9366, 2021.

\bibitem{cordonnier2020relationship}
Jean-Baptiste Cordonnier, Andreas Loukas, and Martin Jaggi.
\newblock On the relationship between self-attention and convolutional layers, 2020.

\bibitem{cubuk2019randaugment}
Ekin~D. Cubuk, Barret Zoph, Jonathon Shlens, and Quoc~V. Le.
\newblock Randaugment: Practical automated data augmentation with a reduced search space, 2019.

\bibitem{dai2017deformable}
Jifeng Dai, Haozhi Qi, Yuwen Xiong, Yi Li, Guodong Zhang, Han Hu, and Yichen Wei.
\newblock Deformable convolutional networks.
\newblock In {\em Proceedings of the IEEE international conference on computer vision}, pages 764--773, 2017.

\bibitem{devlin2019bert}
Jacob Devlin, Ming-Wei Chang, Kenton Lee, and Kristina Toutanova.
\newblock Bert: Pre-training of deep bidirectional transformers for language understanding, 2019.

\bibitem{ding2021repmlp}
Xiaohan Ding, Chunlong Xia, Xiangyu Zhang, Xiaojie Chu, Jungong Han, and Guiguang Ding.
\newblock Repmlp: Re-parameterizing convolutions into fully-connected layers for image recognition.
\newblock {\em arXiv preprint arXiv:2105.01883}, 2021.

\bibitem{dosovitskiy2021image}
Alexey Dosovitskiy, Lucas Beyer, Alexander Kolesnikov, Dirk Weissenborn, Xiaohua Zhai, Thomas Unterthiner, Mostafa Dehghani, Matthias Minderer, Georg Heigold, Sylvain Gelly, Jakob Uszkoreit, and Neil Houlsby.
\newblock An image is worth 16x16 words: Transformers for image recognition at scale, 2021.

\bibitem{fu2024monarch}
Dan Fu, Simran Arora, Jessica Grogan, Isys Johnson, Evan~Sabri Eyuboglu, Armin Thomas, Benjamin Spector, Michael Poli, Atri Rudra, and Christopher R{\'e}.
\newblock Monarch mixer: A simple sub-quadratic gemm-based architecture.
\newblock {\em Advances in Neural Information Processing Systems}, 36, 2024.

\bibitem{pmlr-v9-glorot10a}
Xavier Glorot and Yoshua Bengio.
\newblock Understanding the difficulty of training deep feedforward neural networks.
\newblock In Yee~Whye Teh and Mike Titterington, editors, {\em Proceedings of the Thirteenth International Conference on Artificial Intelligence and Statistics}, volume~9 of {\em Proceedings of Machine Learning Research}, pages 249--256, Chia Laguna Resort, Sardinia, Italy, 13--15 May 2010. PMLR.

\bibitem{go2023gswin}
Mocho Go and Hideyuki Tachibana.
\newblock gswin: Gated mlp vision model with hierarchical structure of shifted window.
\newblock In {\em ICASSP 2023-2023 IEEE International Conference on Acoustics, Speech and Signal Processing (ICASSP)}, pages 1--5. IEEE, 2023.

\bibitem{guo2021hiremlp}
Jianyuan Guo, Yehui Tang, Kai Han, Xinghao Chen, Han Wu, Chao Xu, Chang Xu, and Yunhe Wang.
\newblock Hire-mlp: Vision mlp via hierarchical rearrangement, 2021.

\bibitem{he2018mask}
Kaiming He, Georgia Gkioxari, Piotr Dollár, and Ross Girshick.
\newblock Mask r-cnn, 2018.

\bibitem{he2015deep}
Kaiming He, Xiangyu Zhang, Shaoqing Ren, and Jian Sun.
\newblock Deep residual learning for image recognition, 2015.

\bibitem{hendrycks2023gaussian}
Dan Hendrycks and Kevin Gimpel.
\newblock Gaussian error linear units (gelus), 2023.

\bibitem{heo2021rethinking}
Byeongho Heo, Sangdoo Yun, Dongyoon Han, Sanghyuk Chun, Junsuk Choe, and Seong~Joon Oh.
\newblock Rethinking spatial dimensions of vision transformers.
\newblock In {\em Proceedings of the IEEE/CVF International Conference on Computer Vision}, pages 11936--11945, 2021.

\bibitem{hou2021vision}
Qibin Hou, Zihang Jiang, Li Yuan, Ming-Ming Cheng, Shuicheng Yan, and Jiashi Feng.
\newblock Vision permutator: A permutable mlp-like architecture for visual recognition, 2021.

\bibitem{huang2016deep}
Gao Huang, Yu Sun, Zhuang Liu, Daniel Sedra, and Kilian Weinberger.
\newblock Deep networks with stochastic depth, 2016.

\bibitem{huang2023adaptive}
Zhipeng Huang, Zhizheng Zhang, Cuiling Lan, Zheng-Jun Zha, Yan Lu, and Baining Guo.
\newblock Adaptive frequency filters as efficient global token mixers.
\newblock In {\em Proceedings of the IEEE/CVF International Conference on Computer Vision}, pages 6049--6059, 2023.

\bibitem{kirillov2019panoptic}
Alexander Kirillov, Ross Girshick, Kaiming He, and Piotr Dollár.
\newblock Panoptic feature pyramid networks, 2019.

\bibitem{kolesnikov2020big}
Alexander Kolesnikov, Lucas Beyer, Xiaohua Zhai, Joan Puigcerver, Jessica Yung, Sylvain Gelly, and Neil Houlsby.
\newblock Big transfer (bit): General visual representation learning, 2020.

\bibitem{koohpayegani2024sima}
Soroush~Abbasi Koohpayegani and Hamed Pirsiavash.
\newblock Sima: Simple softmax-free attention for vision transformers.
\newblock In {\em Proceedings of the IEEE/CVF Winter Conference on Applications of Computer Vision}, pages 2607--2617, 2024.

\bibitem{Krizhevsky09learningmultiple}
Alex Krizhevsky.
\newblock Learning multiple layers of features from tiny images.
\newblock Technical report, University of Toronto, 2009.

\bibitem{NIPS2012_c399862d}
Alex Krizhevsky, Ilya Sutskever, and Geoffrey~E Hinton.
\newblock Imagenet classification with deep convolutional neural networks.
\newblock In F. Pereira, C.J. Burges, L. Bottou, and K.Q. Weinberger, editors, {\em Advances in Neural Information Processing Systems}, volume~25. Curran Associates, Inc., 2012.

\bibitem{cnn-lecun}
Y. Lecun, L. Bottou, Y. Bengio, and P. Haffner.
\newblock Gradient-based learning applied to document recognition.
\newblock {\em Proceedings of the IEEE}, 86(11):2278--2324, 1998.

\bibitem{leethorp2022fnet}
James Lee-Thorp, Joshua Ainslie, Ilya Eckstein, and Santiago Ontanon.
\newblock Fnet: Mixing tokens with fourier transforms, 2022.

\bibitem{li2024noah}
Chao Li, Aojun Zhou, and Anbang Yao.
\newblock Noah: Learning pairwise object category attentions for image classification.
\newblock {\em arXiv preprint arXiv:2402.02377}, 2024.

\bibitem{li2023convmlp}
Jiachen Li, Ali Hassani, Steven Walton, and Humphrey Shi.
\newblock Convmlp: Hierarchical convolutional mlps for vision.
\newblock In {\em Proceedings of the IEEE/CVF Conference on Computer Vision and Pattern Recognition}, pages 6306--6315, 2023.

\bibitem{li2022efficientformer}
Yanyu Li, Geng Yuan, Yang Wen, Ju Hu, Georgios Evangelidis, Sergey Tulyakov, Yanzhi Wang, and Jian Ren.
\newblock Efficientformer: Vision transformers at mobilenet speed.
\newblock {\em Advances in Neural Information Processing Systems}, 35:12934--12949, 2022.

\bibitem{lian2022asmlp}
Dongze Lian, Zehao Yu, Xing Sun, and Shenghua Gao.
\newblock As-mlp: An axial shifted mlp architecture for vision, 2022.

\bibitem{lin2018focal}
Tsung-Yi Lin, Priya Goyal, Ross Girshick, Kaiming He, and Piotr Dollár.
\newblock Focal loss for dense object detection, 2018.

\bibitem{lin2015microsoft}
Tsung-Yi Lin, Michael Maire, Serge Belongie, Lubomir Bourdev, Ross Girshick, James Hays, Pietro Perona, Deva Ramanan, C.~Lawrence Zitnick, and Piotr Dollár.
\newblock Microsoft coco: Common objects in context, 2015.

\bibitem{liu2021pay}
Hanxiao Liu, Zihang Dai, David~R. So, and Quoc~V. Le.
\newblock Pay attention to mlps, 2021.

\bibitem{liu2019roberta}
Yinhan Liu, Myle Ott, Naman Goyal, Jingfei Du, Mandar Joshi, Danqi Chen, Omer Levy, Mike Lewis, Luke Zettlemoyer, and Veselin Stoyanov.
\newblock Roberta: A robustly optimized bert pretraining approach, 2019.

\bibitem{liu2022swin}
Ze Liu, Han Hu, Yutong Lin, Zhuliang Yao, Zhenda Xie, Yixuan Wei, Jia Ning, Yue Cao, Zheng Zhang, Li Dong, et~al.
\newblock Swin transformer v2: Scaling up capacity and resolution.
\newblock In {\em Proceedings of the IEEE/CVF conference on computer vision and pattern recognition}, pages 12009--12019, 2022.

\bibitem{liu2021swin}
Ze Liu, Yutong Lin, Yue Cao, Han Hu, Yixuan Wei, Zheng Zhang, Stephen Lin, and Baining Guo.
\newblock Swin transformer: Hierarchical vision transformer using shifted windows, 2021.

\bibitem{liu2022convnet}
Zhuang Liu, Hanzi Mao, Chao-Yuan Wu, Christoph Feichtenhofer, Trevor Darrell, and Saining Xie.
\newblock A convnet for the 2020s.
\newblock In {\em Proceedings of the IEEE/CVF conference on computer vision and pattern recognition}, pages 11976--11986, 2022.

\bibitem{loshchilov2019decoupled}
Ilya Loshchilov and Frank Hutter.
\newblock Decoupled weight decay regularization, 2019.

\bibitem{openai2023gpt4}
OpenAI.
\newblock Gpt-4 technical report, 2023.

\bibitem{poli2023hyena}
Michael Poli, Stefano Massaroli, Eric Nguyen, Daniel~Y Fu, Tri Dao, Stephen Baccus, Yoshua Bengio, Stefano Ermon, and Christopher R{\'e}.
\newblock Hyena hierarchy: Towards larger convolutional language models.
\newblock {\em arXiv preprint arXiv:2302.10866}, 2023.

\bibitem{qin2024hierarchically}
Zhen Qin, Songlin Yang, and Yiran Zhong.
\newblock Hierarchically gated recurrent neural network for sequence modeling.
\newblock {\em Advances in Neural Information Processing Systems}, 36, 2024.

\bibitem{radford2019language}
Alec Radford, Jeff Wu, Rewon Child, David Luan, Dario Amodei, and Ilya Sutskever.
\newblock Language models are unsupervised multitask learners, 2019.

\bibitem{raffel2020exploring}
Colin Raffel, Noam Shazeer, Adam Roberts, Katherine Lee, Sharan Narang, Michael Matena, Yanqi Zhou, Wei Li, and Peter~J. Liu.
\newblock Exploring the limits of transfer learning with a unified text-to-text transformer, 2020.

\bibitem{russakovsky2015imagenet}
Olga Russakovsky, Jia Deng, Hao Su, Jonathan Krause, Sanjeev Satheesh, Sean Ma, Zhiheng Huang, Andrej Karpathy, Aditya Khosla, Michael Bernstein, Alexander~C. Berg, and Li Fei-Fei.
\newblock Imagenet large scale visual recognition challenge, 2015.

\bibitem{shen2023deepmad}
Xuan Shen, Yaohua Wang, Ming Lin, Yilun Huang, Hao Tang, Xiuyu Sun, and Yanzhi Wang.
\newblock Deepmad: Mathematical architecture design for deep convolutional neural network.
\newblock In {\em Proceedings of the IEEE/CVF Conference on Computer Vision and Pattern Recognition}, pages 6163--6173, 2023.

\bibitem{simonyan2015deep}
Karen Simonyan and Andrew Zisserman.
\newblock Very deep convolutional networks for large-scale image recognition, 2015.

\bibitem{sun2017revisiting}
Chen Sun, Abhinav Shrivastava, Saurabh Singh, and Abhinav Gupta.
\newblock Revisiting unreasonable effectiveness of data in deep learning era, 2017.

\bibitem{sun2019ernie}
Yu Sun, Shuohuan Wang, Yukun Li, Shikun Feng, Xuyi Chen, Han Zhang, Xin Tian, Danxiang Zhu, Hao Tian, and Hua Wu.
\newblock Ernie: Enhanced representation through knowledge integration, 2019.

\bibitem{szegedy2014going}
Christian Szegedy, Wei Liu, Yangqing Jia, Pierre Sermanet, Scott Reed, Dragomir Anguelov, Dumitru Erhan, Vincent Vanhoucke, and Andrew Rabinovich.
\newblock Going deeper with convolutions, 2014.

\bibitem{tan2020efficientnet}
Mingxing Tan and Quoc~V. Le.
\newblock Efficientnet: Rethinking model scaling for convolutional neural networks, 2020.

\bibitem{tang2022sparse}
Chuanxin Tang, Yucheng Zhao, Guangting Wang, Chong Luo, Wenxuan Xie, and Wenjun Zeng.
\newblock Sparse mlp for image recognition: Is self-attention really necessary?, 2022.

\bibitem{tang2022image}
Yehui Tang, Kai Han, Jianyuan Guo, Chang Xu, Yanxi Li, Chao Xu, and Yunhe Wang.
\newblock An image patch is a wave: Phase-aware vision mlp, 2022.

\bibitem{tolstikhin2021mlpmixer}
Ilya Tolstikhin, Neil Houlsby, Alexander Kolesnikov, Lucas Beyer, Xiaohua Zhai, Thomas Unterthiner, Jessica Yung, Andreas Steiner, Daniel Keysers, Jakob Uszkoreit, Mario Lucic, and Alexey Dosovitskiy.
\newblock Mlp-mixer: An all-mlp architecture for vision, 2021.

\bibitem{touvron2021resmlp}
Hugo Touvron, Piotr Bojanowski, Mathilde Caron, Matthieu Cord, Alaaeldin El-Nouby, Edouard Grave, Gautier Izacard, Armand Joulin, Gabriel Synnaeve, Jakob Verbeek, and Hervé Jégou.
\newblock Resmlp: Feedforward networks for image classification with data-efficient training, 2021.

\bibitem{touvron2021training}
Hugo Touvron, Matthieu Cord, Matthijs Douze, Francisco Massa, Alexandre Sablayrolles, and Hervé Jégou.
\newblock Training data-efficient image transformers \& distillation through attention, 2021.

\bibitem{touvron2021going}
Hugo Touvron, Matthieu Cord, Alexandre Sablayrolles, Gabriel Synnaeve, and Hervé Jégou.
\newblock Going deeper with image transformers, 2021.

\bibitem{touvron2023llama}
Hugo Touvron, Thibaut Lavril, Gautier Izacard, Xavier Martinet, Marie-Anne Lachaux, Timothée Lacroix, Baptiste Rozière, Naman Goyal, Eric Hambro, Faisal Azhar, Aurelien Rodriguez, Armand Joulin, Edouard Grave, and Guillaume Lample.
\newblock Llama: Open and efficient foundation language models, 2023.

\bibitem{touvron2023llama2}
Hugo Touvron, Louis Martin, Kevin Stone, Peter Albert, Amjad Almahairi, Yasmine Babaei, Nikolay Bashlykov, Soumya Batra, Prajjwal Bhargava, Shruti Bhosale, Dan Bikel, Lukas Blecher, Cristian~Canton Ferrer, Moya Chen, Guillem Cucurull, David Esiobu, Jude Fernandes, Jeremy Fu, Wenyin Fu, Brian Fuller, Cynthia Gao, Vedanuj Goswami, Naman Goyal, Anthony Hartshorn, Saghar Hosseini, Rui Hou, Hakan Inan, Marcin Kardas, Viktor Kerkez, Madian Khabsa, Isabel Kloumann, Artem Korenev, Punit~Singh Koura, Marie-Anne Lachaux, Thibaut Lavril, Jenya Lee, Diana Liskovich, Yinghai Lu, Yuning Mao, Xavier Martinet, Todor Mihaylov, Pushkar Mishra, Igor Molybog, Yixin Nie, Andrew Poulton, Jeremy Reizenstein, Rashi Rungta, Kalyan Saladi, Alan Schelten, Ruan Silva, Eric~Michael Smith, Ranjan Subramanian, Xiaoqing~Ellen Tan, Binh Tang, Ross Taylor, Adina Williams, Jian~Xiang Kuan, Puxin Xu, Zheng Yan, Iliyan Zarov, Yuchen Zhang, Angela Fan, Melanie Kambadur, Sharan Narang, Aurelien Rodriguez, Robert Stojnic, Sergey Edunov, and Thomas
  Scialom.
\newblock Llama 2: Open foundation and fine-tuned chat models, 2023.

\bibitem{trockman2022patches}
Asher Trockman and J~Zico Kolter.
\newblock Patches are all you need?
\newblock {\em arXiv preprint arXiv:2201.09792}, 2022.

\bibitem{vaswani2023attention}
Ashish Vaswani, Noam Shazeer, Niki Parmar, Jakob Uszkoreit, Llion Jones, Aidan~N. Gomez, Lukasz Kaiser, and Illia Polosukhin.
\newblock Attention is all you need, 2023.

\bibitem{wang2021pyramid}
Wenhai Wang, Enze Xie, Xiang Li, Deng-Ping Fan, Kaitao Song, Ding Liang, Tong Lu, Ping Luo, and Ling Shao.
\newblock Pyramid vision transformer: A versatile backbone for dense prediction without convolutions.
\newblock In {\em Proceedings of the IEEE/CVF international conference on computer vision}, pages 568--578, 2021.

\bibitem{wang2022pvt}
Wenhai Wang, Enze Xie, Xiang Li, Deng-Ping Fan, Kaitao Song, Ding Liang, Tong Lu, Ping Luo, and Ling Shao.
\newblock Pvt v2: Improved baselines with pyramid vision transformer.
\newblock {\em Computational Visual Media}, 8(3):415--424, 2022.

\bibitem{wei2022active}
Guoqiang Wei, Zhizheng Zhang, Cuiling Lan, Yan Lu, and Zhibo Chen.
\newblock Active token mixer, 2022.

\bibitem{xiao2018unified}
Tete Xiao, Yingcheng Liu, Bolei Zhou, Yuning Jiang, and Jian Sun.
\newblock Unified perceptual parsing for scene understanding, 2018.

\bibitem{xie2017aggregated}
Saining Xie, Ross Girshick, Piotr Doll{\'a}r, Zhuowen Tu, and Kaiming He.
\newblock Aggregated residual transformations for deep neural networks.
\newblock In {\em Proceedings of the IEEE conference on computer vision and pattern recognition}, pages 1492--1500, 2017.

\bibitem{yang2022focal}
Jianwei Yang, Chunyuan Li, Xiyang Dai, and Jianfeng Gao.
\newblock Focal modulation networks.
\newblock {\em Advances in Neural Information Processing Systems}, 35:4203--4217, 2022.

\bibitem{yang2021focal}
Jianwei Yang, Chunyuan Li, Pengchuan Zhang, Xiyang Dai, Bin Xiao, Lu Yuan, and Jianfeng Gao.
\newblock Focal self-attention for local-global interactions in vision transformers.
\newblock {\em arXiv preprint arXiv:2107.00641}, 2021.

\bibitem{yang2020xlnet}
Zhilin Yang, Zihang Dai, Yiming Yang, Jaime Carbonell, Ruslan Salakhutdinov, and Quoc~V. Le.
\newblock Xlnet: Generalized autoregressive pretraining for language understanding, 2020.

\bibitem{yeh2023attentionviz}
Catherine Yeh, Yida Chen, Aoyu Wu, Cynthia Chen, Fernanda Viégas, and Martin Wattenberg.
\newblock Attentionviz: A global view of transformer attention, 2023.

\bibitem{yu2021s2mlp}
Tan Yu, Xu Li, Yunfeng Cai, Mingming Sun, and Ping Li.
\newblock S$^2$-mlp: Spatial-shift mlp architecture for vision, 2021.

\bibitem{yu2024white}
Yaodong Yu, Sam Buchanan, Druv Pai, Tianzhe Chu, Ziyang Wu, Shengbang Tong, Benjamin Haeffele, and Yi Ma.
\newblock White-box transformers via sparse rate reduction.
\newblock {\em Advances in Neural Information Processing Systems}, 36, 2024.

\bibitem{yun2019cutmix}
Sangdoo Yun, Dongyoon Han, Seong~Joon Oh, Sanghyuk Chun, Junsuk Choe, and Youngjoon Yoo.
\newblock Cutmix: Regularization strategy to train strong classifiers with localizable features, 2019.

\bibitem{zhang2022morphmlp}
David~Junhao Zhang, Kunchang Li, Yali Wang, Yunpeng Chen, Shashwat Chandra, Yu Qiao, Luoqi Liu, and Mike~Zheng Shou.
\newblock Morphmlp: An efficient mlp-like backbone for spatial-temporal representation learning, 2022.

\bibitem{zhang2018mixup}
Hongyi Zhang, Moustapha Cisse, Yann~N. Dauphin, and David Lopez-Paz.
\newblock mixup: Beyond empirical risk minimization, 2018.

\bibitem{zhong2017random}
Zhun Zhong, Liang Zheng, Guoliang Kang, Shaozi Li, and Yi Yang.
\newblock Random erasing data augmentation, 2017.

\bibitem{zhou2018semantic}
Bolei Zhou, Hang Zhao, Xavier Puig, Tete Xiao, Sanja Fidler, Adela Barriuso, and Antonio Torralba.
\newblock Semantic understanding of scenes through the ade20k dataset, 2018.

\bibitem{zhu2024vision}
Lianghui Zhu, Bencheng Liao, Qian Zhang, Xinlong Wang, Wenyu Liu, and Xinggang Wang.
\newblock Vision mamba: Efficient visual representation learning with bidirectional state space model.
\newblock {\em arXiv preprint arXiv:2401.09417}, 2024.

\bibitem{zoph2018learning}
Barret Zoph, Vijay Vasudevan, Jonathon Shlens, and Quoc~V. Le.
\newblock Learning transferable architectures for scalable image recognition, 2018.

\end{thebibliography}
}

\clearpage
\appendix
\section{Related Works}
\subsection{CNN-Based}
For a long range of time, CNN-based architectures have dominated the computer vision domain. The prototype of CNN is presented in~\cite{cnn-lecun} and after the exciting success of AlexNet~\cite{NIPS2012_c399862d}, a large number of methods adopt CNN architecture for higher performance~\cite{simonyan2015deep, szegedy2014going, he2015deep}. Especially the ResNet~\cite{he2015deep}, utilizes a residual connection among layers to alleviate the gradient vanishing. Owing to the hierarchical structure employed in CNNs, rich information can be effectively extracted from localized receptive fields. Yet, despite advantages, CNNs exhibit certain limitations, particularly in terms of capturing global contextual information, and inductive bias both could potentially impede CNNs being applied to downstream tasks.




\subsection{Transformer-Based}
The foundation work~\cite{vaswani2023attention} on Transformer-based architecture proposes the attention mechanism used to extract the relationship among the spatial positional features.  Following the groundbreaking achievements of BERT~\cite{devlin2019bert} in natural language processing, numerous approaches have leveraged Transformer-based architecture for advanced performance~\cite{liu2019roberta, sun2019ernie, raffel2020exploring, yang2020xlnet, radford2019language, brown2020language, openai2023gpt4, touvron2023llama, touvron2023llama2, dosovitskiy2021image, bao2022beit, liu2021swin}, both in natural language processing and computer vision. Notably with self-attention mechanism, Vision Transformer~\cite{dosovitskiy2021image} has facilitated long-range dependencies, indicating that Transformers-based architectures have the ability to extract global context understanding. 



Despite remarkable achievements across various domains, Transformers do have certain drawbacks that need to be considered. $\text{SoftMax}(\frac{Q\times K^T}{\sqrt{k}})V$, is the primary reason for heavy computational demands. Furthermore, the scalability is not only impeded by its quadratic computational complexity, but also by the necessity for extensive datasets and significant memory consumption. 

\section{Compare SpiralMLP to CNNs and MHSA}
\label{sec:appx:1}
In order to establish a comparison between \textbf{Multi-Head Self-Attention} (abbreviated as \textbf{MHSA}) and \textbf{SpiralMLP}, it is necessary to demonstrate the relationship between MHSA and \textbf{Convolutional Neural Networks} (abbreviated as \textbf{CNNs}). This is because SpiralMLP, unlike MHSA, does not incorporate self-attention layers and is more closely aligned with CNNs.

By examining the connection between MHSA and CNNs, we can provide a comprehensive understanding of the architectural differences and similarities between MHSA and SpiralMLP. This will enable us to highlight the unique features and advantages of each approach. 

Thus we outline the following subsections in this way: 
\begin{itemize}
\item   We first elaborate how SpiralMLP is related to CNNs.
\item   Then we draw a relation between MHSA and CNNs using the proof of Cordonnier~\cite{cordonnier2020relationship}.
\item   Finally we provide the comparison between MHSA and SpiralMLP.
\end{itemize}

\subsection{How is SpiralMLP related to CNNs?}
\label{sec:appx:spiral-cnn}


In order to demonstrate the functionality of CNNs, we define the standard convolution weight matrix as $W^{\text{cnn}}\in\mathbb{R}^{K_H\times K_W\times C_{\text{in}}\times C_{\text{out}}}$, where $K_H$, $K_W$, $C_{\text{in}}$, $C_{\text{out}}$ are the kernel height, kernel width, input channel dimension, output channel dimension, respectively. Given a position $(i,j,:)$ within the feature map $X\in\mathbb{R}^{H\times W\times C_{\text{in}}}$, where $H$, $W$ are the height and weight, and $\Omega=\{(x,y);0\le x\le K_W,0\le y\le K_H\}$ is defined as a set of coordinates in a rectangular, corresponding to the kernel. Thus the convolution operation is formulated as:

\begin{align}
\text{Conv}_{i,j,:}(X)=\sum_{(x,y)\in\Omega}X_{i+x,j+y,:}W^{\text{cnn}}_{y,x,:,:}+b^{\text{cnn}}
\label{appxeq:1}
\end{align}

where, $b^{\text{cnn}}\in\mathbb{R}^{C_{\text{out}}}$ is the CNNs bias, the overall output $\text{Conv}_{i,j,\colon}(\cdot)$ is a vector at each position $(i,j,:)$.


Next, we replace the dot product between $X_{i+x, j+y,:}$ and $W^{\text{cnn}}_{y,x,:,:}$ in~\cref{appxeq:1} with a summation across the channel dimension to get:

\begin{align}
    \text{Conv}_{i,j,:}(X) = \sum_{(x,y)\in\Omega}\sum^{C_{\text{in}}}_{c=0}X_{i+x,j+y,c}W^{\text{cnn}}_{y,x,c,:}+b^{\text{cnn}}\label{appxeq:2}
\end{align}

Afterwards we change the order of the summation and obtain: 

\begin{align}
\text{Conv}_{i,j,:}(X)=\sum^{C_{\text{in}}}_{c=0}\sum_{(x,y)\in\Omega}X_{i+x,j+y,c}W^{\text{cnn}}_{y,x,c,:}+b^{\text{cnn}}\label{appxeq:3}
\end{align}

To relate~\cref{appxeq:3} to the Spiral FC (\cref{eq:5,eq:6,eq:7}), we first create a function:
\begin{align}
g(x,y,c)=\begin{cases}1,\quad&\text{if}\ x=\phi_i(c)\ \text{and}\ y=\phi_j(c)\\ 0,\quad&\text{otherwise} \end{cases}\label{appxeq:4}
\end{align}

Then we apply the~\cref{appxeq:4} on the kernel weights $W^{\text{cnn}}$ of the CNNs to obtain: 
\begin{align}
\widetilde{W}^{\text{cnn}}_{y,x,c,:}=g(x,y,c)\cdot W^{\text{cnn}}_{y,x,c,:}\label{appxeq:5}
\end{align}

Subsequently, we substitute the weights defined in~\cref{appxeq:3} with the weights in~\cref{appxeq:5}:
\begin{equation}
\begin{aligned}
&\text{Conv}_{i,j,:}(X)\\
&=\sum^{C_{\text{in}}}_{c=0}\sum_{(x,y)\in\Omega}X_{i+x,j+y,c}\widetilde{W}^{\text{cnn}}_{y,x,c,:}+b^{\text{cnn}} \\ 
&=\sum^{C_{\text{in}}}_{c=0}\sum_{(x,y)\in\Omega}g(x,y,c)X_{i+x,j+y,c}W^{\text{cnn}}_{y,x,c,:}+b^{\text{cnn}}\\ 
&=\sum^{C_{\text{in}}}_{c=0}X_{i+\phi_i(c),j+\phi_j(c),c}W^{\text{spiral}}_{c,:}+b^{\text{spiral}}\\ 
&=\text{Spiral FC}_{i,j,:}(X)
\end{aligned}\label{appxeq:6}
\end{equation}

where, $W^{\text{spiral}}$ and $b^{\text{spiral}}$ are already introduced in~\cref{eq:5}.

Consequently, we have demonstrated how the Spiral FC can be derived from convolutions. It is noteworthy that~\cref{appxeq:4,appxeq:5,appxeq:6} reveal that the Spiral FC assumes sparsity within the convolution operation, indicating its potential advantages and distinctive characteristics compared to traditional dense convolution layers.

\subsection{How is MHSA related to CNNs?}
\label{sec:appx:mhsa-cnn}

To formulate, the multi-head self-attention weight matrix for each head is denoted as $W^{\text{mhsa,}h}\in\mathbb{R}^{C_{\text{in}}\times C_{\text{out}}}$, and the bias is denoted as $b^{\text{mhsa}}\in\mathbb{R}^{C_{\text{out}}}$. Furthermore, $h$ refers to each individual head and $N_h$ represents the total number of heads. The process of MHSA applied to the input $X\in\mathbb{R}^{H\times W\times C_{\text{in}}}$ on position $(i,j,:)$ can be arranged as:
\begin{equation}
    \text{MHSA}_{i,j,:}(X)=\sum^{N_h}_{h=1}X_{:,:,C^{h}_{\text{in}}}W^{\text{mhsa},h}_{C^{h}_{\text{in}},:}+b^{\text{mhsa}}
    \label{appxeq:attn}
\end{equation}

where, $C^h_{\text{in}}$ illustrates the division of the input channel corresponding to the $h^{\text{th}}$ head, and it is incorporated into the computation for each head $h$. 

While in the work~\cite{cordonnier2020relationship}, it is demonstrated that MHSA can be represented in a manner similar to CNNs: 

\begin{equation}
\begin{aligned}
&\text{MHSA}_{i,j,:}(X)\\
&= \sum^{N_h}_{h=1}X_{i+\Delta_i(h),j+\Delta_j(h),C^h_{\text{in}}}W^{\text{mhsa},h}_{C^h_{\text{in}},:}+b^{\text{mhsa}}\\ 
&= \sum^{N_h}_{h=1}\sum^{C^h_{\text{in}}}_{c=0}X_{i+\Delta_i(h),j+\Delta_j(h),c}W^{\text{mhsa},h}_{c,:}+b^{\text{mhsa}}
\end{aligned}\label{appxeq:7}
\end{equation}

where, all relative positional shifts for a kernel of size $\sqrt{N_h}\times\sqrt{N_h}$ at position $(i,j)$ are contained in $\{\Delta_i(h),\Delta_j(h)\}=\{(-1,0),(-1,1),(-1,2),\dots\}$.

\subsection{How is SpiralMLP related to MHSA?}
Through a similar technique to~\cref{appxeq:5}, we establish a connection between MHSA and SpiralMLP through CNNs. Firstly, we augment $W^{\text{mhsa},h}$ by multiplying with the~\cref{appxeq:4} to obtain: 
\begin{align}
\widetilde{W}^{\text{mhsa},h}_{c,:}=g(x,y,c)\cdot W^{\text{mhsa},h}_{c,:}\label{appxeq:8}
\end{align}

where, $\widetilde{W}^{\text{mhsa},h}$ is dependent on the position along the kernel height, width and input channel. Substitute it into~\cref{appxeq:7}: 
\begin{equation}
\begin{aligned}
&\text{MHSA}_{i,j,:}(X)\\
&=\sum^{N_h}_{h=1}\sum^{C^h_{\text{in}}}_{c=0}X_{i+\Delta_i(h), j+\Delta_j(h),c}\widetilde{W}^{\text{mhsa},h}_{c,:}+b^{\text{mhsa}}\\
&=\sum^{N_h}_{h=1}\sum^{C^h_{\text{in}}}_{c=0}X_{i+\phi_i(c), j+\phi_j(c), c}W^{\text{mhsa},h}_{c,:}+b^{\text{mhsa}}\\
&=\sum^{C_{\text{in}}}_{c=0}X_{i+\phi_i(c), j+\phi_j(c), c}W^{\text{spiral}}_{c,:}+b^{\text{spiral}}\\
&=\text{Spiral FC}_{i,j,:}(X)
\end{aligned}\label{appxeq:9}
\end{equation}

where, $W^{\text{spiral}}$ and $b^{\text{spiral}}$ are already introduced in~\cref{eq:5}.

~\cref{appxeq:9} demonstrates the relationship between Spiral FC and MHSA. Drawing a parallel with the relationship between CNNs and Spiral FC, we can conclude that Spiral FC exhibits a significantly sparser receptive field compared to MHSA. This highlights the distinctive characteristic of Spiral FC in terms of its sparse attention mechanism compared to the dense attention mechanism of MHSA.


\begin{table*}[t]
    \centering
    \begin{tabular}{c|c|c|c|c|c|c|c}
        \hline
        \multirow{2}{*}{\empty} & \multirow{2}{*}{Output Size} & \multirow{2}{*}{Layer Name} & \multicolumn{5}{c}{PVT-Style} \\\cline{4-8}
        &&& B1 & B2 & B3 & B4 & B5\\
        \hline 
        
        \multirow{4}{*}{Stage1} & \multirow{4}{*}{$\frac{H}{4}\times\frac{W}{4}$} & Overlapping & \multicolumn{5}{c}{$S_1=4$}\\\cline{4-8}
        && Patch Embedding & \multicolumn{4}{c|}{$C_1=64$} & $C_1=96$\\\cline{3-8}
        && SpiralMLP & $E_1=4$ & $E_1=4$ & $E_1=4$ & $E_1=4$ & $E_1=4$ \\
        && Block & $L_1=2$ & $L_1=2$ & $L_1=3$ & $L_1=3$ & $L_1=3$ \\\hline

       \multirow{4}{*}{Stage2} & \multirow{4}{*}{$\frac{H}{8}\times\frac{W}{8}$} & Overlapping & \multicolumn{5}{c}{$S_2=2$}\\\cline{4-8}
        && Patch Embedding & \multicolumn{4}{c|}{$C_2=128$} & $C_2=192$\\\cline{3-8}
        && SpiralMLP & $E_2=4$ & $E_2=4$ & $E_2=4$ & $E_2=4$ & $E_2=4$ \\
        && Block & $L_2=2$ & $L_2=3$ & $L_2=4$ & $L_2=8$ & $L_2=4$ \\\hline

        \multirow{4}{*}{Stage3} & \multirow{4}{*}{$\frac{H}{16}\times\frac{W}{16}$} & Overlapping & \multicolumn{5}{c}{$S_3=2$}\\\cline{4-8}
        && Patch Embedding & \multicolumn{4}{c|}{$C_3=320$} & $C_3=384$\\\cline{3-8}
        && SpiralMLP & $E_3=4$ & $E_3=4$ & $E_3=4$ & $E_3=4$ & $E_3=4$ \\
        && Block & $L_3=4$ & $L_3=10$ & $L_3=18$ & $L_3=27$ & $L_3=24$ \\\hline

        \multirow{4}{*}{Stage4} & \multirow{4}{*}{$\frac{H}{32}\times\frac{W}{32}$} & Overlapping & \multicolumn{5}{c}{$S_4=2$}\\\cline{4-8}
        && Patch Embedding & \multicolumn{4}{c|}{$C_4=512$} & $C_4=768$\\\cline{3-8}
        && SpiralMLP & $E_4=4$ & $E_4=4$ & $E_4=4$ & $E_4=4$ & $E_4=4$ \\
        && Block & $L_4=2$ & $L_4=3$ & $L_4=3$ & $L_4=3$ & $L_4=3$ \\\hline

        \multicolumn{3}{c|}{Parameters (M)} & 14 & 24 & 34 & 46 & 68 \\\hline
        \multicolumn{3}{c|}{FLOPs (G)} & 2.0 & 3.6 & 5.6 & 8.2 & 11.3 \\\hline
        \multicolumn{3}{c|}{Accuracy Top-1 (\%)} & 79.8 & 81.9 & 83.4 & 83.8 & 84.0 \\\hline
    \end{tabular}
    \caption{Configurations of SpiralMLP variants in PVT-style. }
    \label{tab:appx:1}
\end{table*}

\section{Model Zoo and Training Details}
\label{sec:appx:5}

\subsection{Model Zoo Configurations and Performances}
\label{sec:appx:model_zoo}
We implement five models according to PVT style, named \textbf{SpiralMLP B1} to \textbf{B5}, and three models based on Swin-style, named as \textbf{SpiralMLP-T}, \textbf{SpiralMLP-S}, \textbf{SpiralMLP-B}. 

SpiralMLP variants in PVT-style are detailed in~\cref{tab:appx:1}. Each stage contains multiple Spiral Blocks with uniform configurations, where the parameters $S$, $E$, $C$, and $L$ denote the shift size, expansion ratio, channel dimension, and the number of layers in each stage, respectively.

\begin{figure*}[tb]
  \centering
    \fbox{\includegraphics[width=0.95\linewidth]{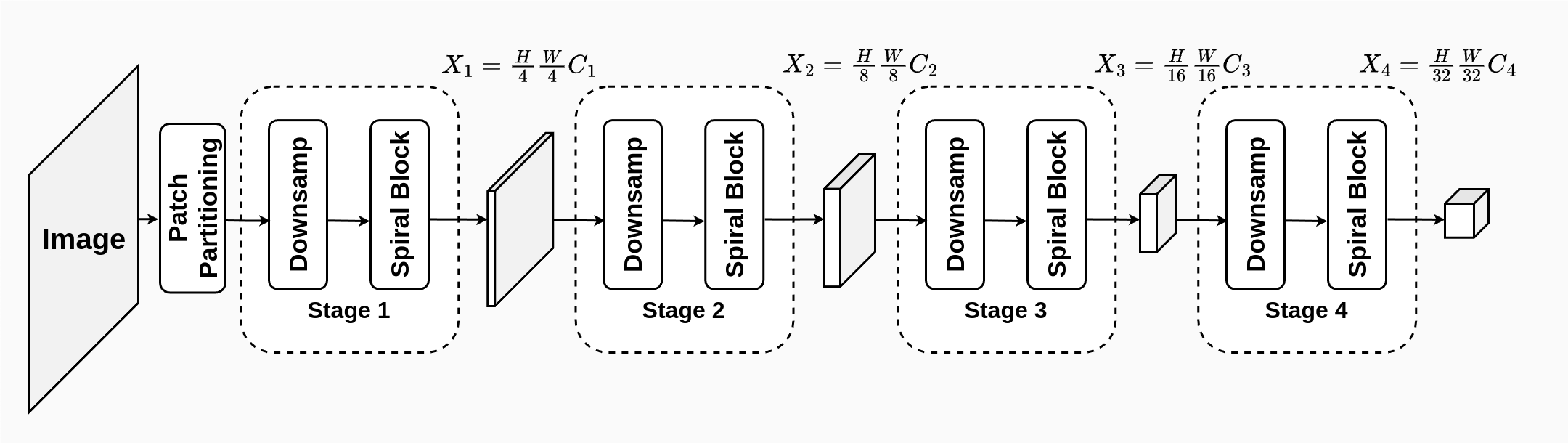}}
    \caption{The architecture of SpiralMLP in Swin-style. }
    \label{fig:appx:swinarch}
\end{figure*}

SpiralMLP variants in Swin-style are shown in~\cref{tab:appx:swin,fig:appx:swinarch}. The input image dimension is $224\times224$. The process 'concat $n\times n$' refers to the concatenation of features from $n\times n$ neighboring features in a patch. This operation effectively downsamples the feature map by a factor of $n$. The notation '96-d' represents a linear layer whose output dimension is 96. In the term [(3,2), 96] indicates that $A_{\text{max}}=3$, $k=2$, $C_{\text{out}}=96$.

\subsection{Experimental Setup for Image Classification}
\label{sec:appx:imgclsset}
We train our model on ImageNet-1k~\cite{russakovsky2015imagenet}, which contains about 1.2M images. The accuracy report is standard Top-1 accuracy on the validation set containing roughly 50k images, evenly distributed among 1000 categories. The code of our implementation is inspired by CycleMLP as well as DeiT and is written in Pytorch. Our augmentation policy includes RandAugment~\cite{cubuk2019randaugment}, Mixup~\cite{zhang2018mixup}, Cutmix~\cite{yun2019cutmix}, random erasing~\cite{zhong2017random} and stochastic depth~\cite{huang2016deep}. The optimizer used is AdamW~\cite{loshchilov2019decoupled} with learning rate of $5\times10\text{e-}4$ with momentum of $0.9$ and weight decay of $5\times10\text{e-}2$.

\subsection{Experimental Setup for Object detection and Instance Segmentation}
\label{sec:appx:objdetset}
For object detection and instance segmentation experiments, we train our model on COCO~\cite{lin2015microsoft} containing $118$k training images together with $5$k validation images. We employ the mmdetection toolbox~\cite{mmdetection} and use RetinaNet~\cite{lin2018focal} and Mask R-CNN~\cite{he2018mask} as the framworks with SpiralMLP variants as backbones. The weights are initialized with the pretrained weights from ImageNet-1k and additional added layers are initialized using Xavier~\cite{pmlr-v9-glorot10a} initialization. The optimizer is AdamW~\cite{loshchilov2019decoupled} with a learning rate of $1\times10\text{e-}4$. The images are resized to $800$ for the shorter side and a maximum limit of $1333$ pixels for height and width of the image. The model is trained on $4$ A100 GPUs with the batch size of $32$ for $12$ epochs.

\subsection{Experimental Setup for Semantic Segmentation}
\label{sec:appx:semsegset}
The semantic segmentation is conducted on ADE20K~\cite{zhou2018semantic}, which consists of 20k training images and 2k validation images. The frameworks used for this purpose are Semantic FPN~\cite{kirillov2019panoptic} and UperNet~\cite{xiao2018unified}, employing SpiralMLP with ImageNet-1k pretrained weights as the backbones. For optimization, the AdamW~\cite{loshchilov2019decoupled} optimizer is chosen. In the case of Mask R-CNN, the optimizer is set with an initial learning rate of $0.0001$ and a weight decay of $0.05$. For Semantic FPN, the same initial learning rate of $0.0001$ is used, but the weight decay is lower, at $0.0001$. The training process is carried out on $4$ A100 GPUs. A batch size of $32$ is maintained throughout, and the model undergoes training for $12$ epochs.

\begin{table*}[tb]
    \centering
    \resizebox{0.95\linewidth}{!}{
    \begin{tabular}{c|c|c|c|c}
        \hline
         & Output Size & SpiralMLP-T & SpiralMLP-S & SpiralMLP-B\\
        \hline 
        
        \multirow{2}{*}{Stage1} & \multirow{2}{*}{$4\times(56\times56)$} & concat $4\times4$, 64-d, LN & concat $4\times4$, 96-d, LN & concat $4\times4$, 96-d, LN \\\cline{3-5}
        && [(3,2), 64]$\times$2 & [(3,2), 96]$\times$3 & [(3,2), 96]$\times$3 \\\hline

        \multirow{2}{*}{Stage2} & \multirow{2}{*}{$8\times(28\times28)$} & concat $2\times2$, 128-d, LN & concat $2\times2$, 192-d, LN & concat $2\times2$, 192-d, LN \\\cline{3-5}
        && [(3,2), 128]$\times$2 & [(3,2), 192]$\times$4 & [(3,2), 192]$\times$4 \\\hline

        \multirow{2}{*}{Stage3} & \multirow{2}{*}{$16\times(14\times14)$} & concat $2\times2$, 320-d, LN & concat $2\times2$, 384-d, LN & concat $2\times2$, 384-d, LN \\\cline{3-5}
        && [(3,2), 320]$\times$6 & [(3,2), 384]$\times$18 & [(3,2), 384]$\times$24 \\\hline
        
        \multirow{2}{*}{Stage4} & \multirow{2}{*}{$32\times(7\times7)$} & concat $2\times2$, 512-d, LN & concat $2\times2$, 768-d, LN & concat $2\times2$, 768-d, LN \\\cline{3-5}
        && [(3,2), 512]$\times$2 & [(3,2), 768]$\times$3 & [(3,2), 768]$\times$3 \\\hline
        
        \multicolumn{2}{c|}{Parameters (M)} & 15 & 56 & 67\\\hline
        \multicolumn{2}{c|}{FLOPs (G)} & 2.3 & 9.1 & 11.0 \\\hline
        \multicolumn{2}{c|}{Accuracy Top-1 (\%)} & 79.6 & 83.3 & 83.6 \\\hline
    \end{tabular}}
    \caption{Configurations of SpiralMLP variants in Swin-style. }
    \label{tab:appx:swin}
\end{table*}

\section{Merge Head}
\subsection{How does the Merge Head Work? }
In~\cref{eq:9}, Merge Head employs a trainable weights $a$ to determine the contribution of $X^{\text{self}}\in\mathbb{R}^{H\times W\times C_{\text{out}}}$, $X^{\text{cross}}\in\mathbb{R}^{H\times W\times C_{\text{out}}}$, from Channel FC and Spiral FC, respectively. For a clearer understanding, we present the tensor shape of each step in Merge Head, as shown in~\cref{tab:appx:mergehead}. Rather than merely employing a straightforward addition, the merge head innovatively takes into account the inputs themselves to formulate the trainable weights $a$. This approach allows for a more dynamic and input-responsive weight adjustment, enhancing the effectiveness of the merging process.

\subsection{Complexity of Merge Head. }
Two inputs are both of shape $\mathbb{R}^{H\times W\times C}$. The addition has $O(HW)$. While the reshaping operation $\mathcal{F}(\cdot)$ is implemented by \texttt{torch.flatten()} with $O(1)$. The average calculation is implemented by \texttt{torch.mean()} with $O(HW)$. The multiplication involves $W^{\text{merge}}\in\mathbb{R}^{2,1}$ contributes $O(1)$. Therefore, the total complexity of this process is linear $O(HW)$.

\begin{table}[tb]
    \centering
    \begin{tabular}{cc}
    \toprule
    Step & Tensor Shape \\
    \midrule
    $(X^{\text{self}},X^{\text{cross}})$ & $(\mathbb{R}^{H\times W\times C_{\text{out}}},\mathbb{R}^{H\times W\times C_{\text{out}}})$ \\
    $\mathcal{F}(\cdot)$ & $\mathbb{R}^{H\cdot W\times C_{\text{out}}}$ \\
    Average & $\mathbb{R}^{1\times C_{\text{out}}}$ \\
    $W^{\text{merge}}$ & $\mathbb{R}^{2\times C_{\text{out}}}$ \\
    \bottomrule
    \end{tabular}
    \caption{Tensor shape of each step in Merge Head. }
    \label{tab:appx:mergehead}
\end{table}


\section{Why Spiral FC works?}

\cref{appxeq:6} indicates that SpiralMLP, along with other criss-cross MLPs, can be effectively implemented using a specialized convolution layer. Currently, deformable convolution~\cite{dai2017deformable} emerges as the optimal method for this purpose. 

In the given scenario as shown in~\cref{fig:appx:explain}, points are identified as $(o_x, o_y)$, $(\mu_x, \mu_y)$, $(\nu_x, \nu_y)$, and $(\mu_x, \nu_y)$, all located on feature map $X$. Specifically, when $(o_x, o_y)$ is considered the reference point, the points $(\mu_x, \mu_y)$ and $(\nu_x, \nu_y)$ are categorized into set $P$. This set includes pairs that are either horizontally or vertically aligned with the reference point $(o_x, o_y)$. The point $(\mu_x, \nu_y)$ falls into set $Q$, which comprises pairs that can be located at any position within the feature map $X$. Then the set $P$ and set $Q$ are defined as:

\begin{figure*}[tb]
  \centering
  \fbox{\includegraphics[width=0.95\linewidth]{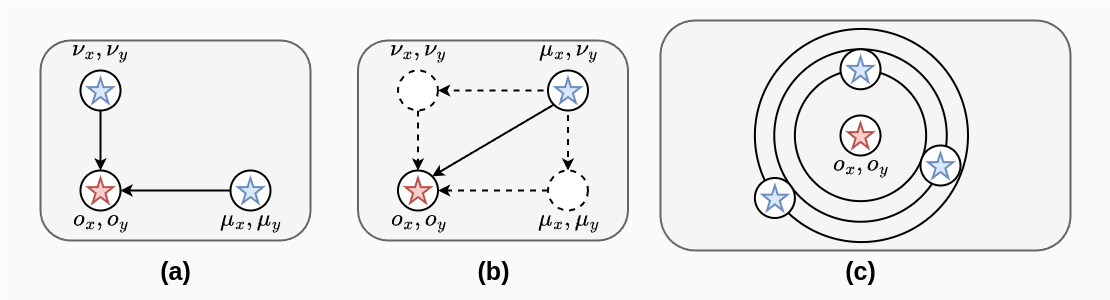}}
   \caption{Spiral FC covers a more comprehensive receptive field. }
   \label{fig:appx:explain}
\end{figure*}

\begin{align}
    P &= \{(\mu_x,\mu_y)\in\mathbb{R}^2|(\mu_x=o_x \lor \mu_y = o_y) \}\label{eq:appx:setp}\\
    Q &= \{(\mu_x,\nu_y)\in\mathbb{R}^2\}\label{eq:appx:setq}
\end{align}

We further denote the $Y_{C}\in\mathbb{R}^{H\times W\times C}$ as the output of the convolution layer, while $X\in\mathbb{R}^{H\times W\times C}$ is still the feature map, this is similar to~\cref{appxeq:6}: 

\begin{equation}
    Y_{C} =\sum^{C_{\text{in}}}_{c=0}X_{i+\bar\phi_i(c),j+\bar\phi_j(c),c}W_{c,:}+b
    \label{appxeq:conv}
\end{equation}

where, $\bar\phi_i(\cdot),\bar\phi_j(\cdot)$ denote the universal offset functions. 


We take two architectures with criss-cross offset functions as examples. When using CyelcMLP~\cite{chen2022cyclemlp} as an example, the offset function is updated into: 
\begin{align}
    \phi^{\text{cycle}}_i(c)&=(c \mod S_H) - 1\label{appxeq:cycle1}\\
    \phi^{\text{cycle}}_j(c)&=(\lfloor\frac{c}{S_H} \rfloor \mod S_W) - 1\label{appxeq:cycle2}
\end{align}

where, $S_H,S_W$ are the predefined step size.  

When using ASMLP~\cite{lian2022asmlp}, the offset function is updated into: 
\begin{align}
    \phi^{\text{as}}_i(c)&=\lfloor\frac{c}{C/s}\rfloor-\lfloor\frac{s}{2}\rfloor\cdot d\label{appxeq:as1}\\
    \phi^{\text{as}}_j(c)&=\lfloor\frac{c}{C/s}\rfloor-\lfloor\frac{s}{2}\rfloor\cdot d\label{appxeq:as2}
\end{align}
where, $s$ is the shift size and $d$ is the dilation rate. 

In~\cref{fig:appx:explain} (a), points \((\mu_x, \mu_y)\) and \((\nu_x, \nu_y)\) from set \(P\) are reachable by the offset functions $\phi^{\text{cycle}}(\cdot)$, and $\phi^{\text{as}}(\cdot)$ due to their placement along the horizontal or vertical axes. However, as illustrated in~\cref{fig:appx:explain} (b), points that do not lie on these axes pose a challenge for the criss-cross methodology, which inherently lacks the capability to capture such spatial information effectively.

To address this limitation,~\cref{fig:appx:explain} (c) suggests the adoption of a predefined multi-helix offset approach. This method, while effective, still offers room for optimization to achieve model efficiency in terms of size and computational speed. A viable solution is the refinement of the multi-helix offset into a spiral-like offset function. This adjustment not only enables the model to recognize points from set $Q$—those not aligned with the horizontal or vertical axes—but also maintains a compact model architecture and ensures rapid processing speeds.


\begin{table}
    \centering
    \resizebox{\linewidth}{!}{
    \begin{tabular}{ccccc}
    \toprule
    Model & Params(M) & Throughput (/sec) & Wall-Clock (s)\\
    \midrule
    Spiral-B1 &  14 &  425 & 22.8 \\
    Spiral-B2 &  24 &  253 & 30.5 \\
    Spiral-B3 &  34 &  143 & 52.2 \\
    Spiral-B4 &  46 &  94 & 72.5 \\
    Spiral-B5 &  68 &  89 & 79.2 \\
    \midrule
    Cycle-B1~\cite{chen2022cyclemlp}  &  15 &  347 & 23.8 \\
    Cycle-B2  &  27 &  207 & 36.0 \\
    Cycle-B3  &  38 &  117 & 60.4 \\
    Cycle-B4  &  52 &  79 & 84.0 \\
    Cycle-B5  &  76 &  75 & 92.0 \\
    ResNet-152~\cite{he2015deep}  & 60 & 152 & 45.7 \\
    Vit-B/16~\cite{dosovitskiy2021image}  & 86 & 102 & 75.2 \\
    Deit-B/16~\cite{touvron2021training} & 86 & 97 & 78.3 \\    
    \bottomrule
    \end{tabular}}
    \caption{Throughput and wall-clock test. }
    \label{tab:appx:throughput}
\end{table}

\section{Latency \& Throughput Analysis}
\label{appx:latency_analysis}
As shown in~\cref{tab:appx:throughput}, throughput testing is conducted on a single NVIDIA A100 GPU, with a batch size of 32 and image resolution of 3x224x224; wall-clock testing is conducted on ImageNet-1k with a single NVIDIA A100 GPU, with a batch size of 64, the number of total batches to 100 and image resolution of 3x224x224. Furthermore, we evaluate the latency speeds on one A100 across various architectures for input resolutions of $224^2$, $384^2$, and $512^2$. These results are detailed in ~\cref{tab:appx:latency}. 

It is evident that SpiralMLP is faster than other MLP-based models when considering the scale of model size. Additionally, the PVT~\cite{wang2021pyramid} showcases even faster processing speeds, mainly due to its utilization of dot product operations, inherently benefiting from accelerated computation. The models used for evaluation are directly extracted from their official implementations. 

\begin{table*}
    \centering
    \resizebox{0.95\linewidth}{!}{
    \begin{tabular}{cccccccccc}
    \toprule
    Model & $224^2$ & $384^2$ & $512^2$ & Params(M) & Model & $224^2$ & $384^2$ & $512^2$ & Params(M) \\
    \midrule
    \cellcolor{LightGray}\textbf{Spiral-B1} & \cellcolor{LightGray}11.74 & \cellcolor{LightGray}11.83 & \cellcolor{LightGray}11.52 & \cellcolor{LightGray}14 & Cycle-B1~\cite{chen2022cyclemlp} & 12.12 & 12.06 & 12.23 & 15 \\ 
    \cellcolor{LightGray}\textbf{Spiral-B2} & \cellcolor{LightGray}20.81 & \cellcolor{LightGray}21.11 & \cellcolor{LightGray}20.29 & \cellcolor{LightGray}24 & Cycle-B2 & 21.66 & 21.36 & 21.12 & 27 \\ 
    \cellcolor{LightGray}\textbf{Spiral-B3} & \cellcolor{LightGray}32.33 & \cellcolor{LightGray}32.53 & \cellcolor{LightGray}31.38 & \cellcolor{LightGray}34 & Cycle-B3 & 33.23 & 32.89 & 32.29 & 38 \\ 
    \cellcolor{LightGray}\textbf{Spiral-B4} & \cellcolor{LightGray}47.00 & \cellcolor{LightGray}47.39 & \cellcolor{LightGray}45.95 & \cellcolor{LightGray}46 & Cycle-B4 & 47.86 & 48.29 & 46.73 & 52 \\ 
    \cellcolor{LightGray}\textbf{Spiral-B5} & \cellcolor{LightGray}39.22 & \cellcolor{LightGray}39.34 & \cellcolor{LightGray}38.34 & \cellcolor{LightGray}68 & Cycle-B5 & 41.00 & 40.87 & 39.59 & 76 \\ 
    \midrule
    ATM-xT~\cite{wei2022active} & 15.01 & 15.32 & 14.86 & 15 & Wave-T-dw~\cite{tang2022image} & 16.91 & 15.83& 16.01 & 15\\
    ATM-T & 25.83 & 26.66 & 25.43 & 27 & Wave-T & 18.56 & 17.48 & 17.47 &17\\
    ATM-S & 37.47 & 38.41 & 37.21 & 39 & Wave-S & 31.97 & 30.76 & 30.61 &31\\
    ATM-B & 55.45 & 55.85 & 53.39 & 52 & Wave-M & 48.98 & 47.43 & 46.94 & 44\\
    ATM-L & 47.24 & 45.98 & 44.78 & 76 & Wave-B & 41.99 & 40.50 & 40.27 & 64\\
    \midrule
    PVT-Tiny~\cite{wang2021pyramid} & 8.54 & 8.72 & 8.55 & 13 & PVTv2-B1~\cite{wang2022pvt} & 9.08 & 9.04 & 9.12 & 14\\
    PVT-Small & 15.73 & 15.75 & 15.62 & 25 & PVTv2-B2 & 17.14 & 17.45 & 17.21 & 25\\
    PVT-Medium & 26.78 & 27.48 & 26.71 & 44 & PVTv2-B3 & 30.57 & 30.21 & 30.32 & 45\\
    PVT-Large & 39.28 & 39.75 & 38.68 & 61 & PVTv2-B4 & 44.77 & 44.73 & 43.51 & 63\\
    -&-&-&-&- & PVTv2-B5 & 56.64 & 56.88 & 55.19 & 82\\
    \bottomrule
    \end{tabular}}
    \caption{Latency analysis measured in \textit{milliseconds(ms)} on one A100. A single image with differing resolutions serves as the input }
    \label{tab:appx:latency}
\end{table*}


\section{Resolution Compatibility}
We compare the image classification models resolution compatibility with Top-1 accuracy on ImageNet-1k, with models pre-trained on 224 x 224 images and tested at various other resolutions without additional fine-tuning, as shown in~\cref{tab:appx:resolution}. 

\begin{table}
    \centering
    \resizebox{0.95\columnwidth}{!}{
    \begin{tabular}{cccccc}
    \toprule
    Model & Params(M) & 128 & 224 & 256 & 384 \\
    \midrule
    Spiral-B5 & 68 & 77.6 & 84.0 & 83.8 & 83.4 \\ 
    Cycle-B~\cite{chen2022cyclemlp} & 88 & 77.0 & 83.2 & 83.1 & 82.7 \\
    ViT-B/16~\cite{dosovitskiy2021image} & 86 & - & 81.0 & 81.6 & 79.0 \\
    ResNet-152~\cite{he2015deep} & 60 & 68.7 & 78.3 & 78.4 & 77.4 \\
    DeiT-B/16~\cite{touvron2021training} & 87 & - & 81.7 & - & 82.5 \\
    
    \bottomrule
    \end{tabular}}
    \caption{Resolution compatibility comparison. }
    \label{tab:appx:resolution}
\end{table}


\section{Spiral in AttentionViz}
In the work AttentionViz~\cite{yeh2023attentionviz}, the authors introduce a visualization tool for analyzing transformer models, particularly focusing on the interaction between keys and queries distributions across various heads and layers. A notable discovery in their research is the identification of a spiral-like pattern. This pattern suggests that the keys and queries in transformers are spatially aligned in a manner resembling a spiral. The phenomenon results from position vectors that are generated using trigonometric functions, mapping onto a helical curve in a high-dimensional space. In linguistics models, this refers to the arrangement of words or parts of words; while in vision transformers, it relates to the organization of pixel patches.

This spiral distribution is reflective of the initial ordering vector given to transformers, illustrating how positional information is embedded within the model. Furthermore, the research reveals that transformer heads displaying a spiral shape or having clusters of queries/keys tend to yield more dispersed search results.

\end{document}